\documentclass[10pt,twocolumn,letterpaper]{article}

\usepackage[pagenumbers]{cvpr} %

\usepackage{colortbl}
\definecolor{mygray}{gray}{.95}
\usepackage{bm}
\usepackage[linesnumbered,ruled,vlined]{algorithm2e}
\usepackage{multirow}

\definecolor{cvprblue}{rgb}{0.21,0.49,0.74}
\usepackage[pagebackref,breaklinks,colorlinks,allcolors=cvprblue]{hyperref}

\def\paperID{11366} %
\def\confName{CVPR}
\def\confYear{2025}

\title{Multi-Task Model Merging via Adaptive Weight Disentanglement}

\author{Feng Xiong\textsuperscript{1\thanks{\quad Equal Contribution.}},
Runxi Cheng\textsuperscript{2$^\ast$},
Wang Chen\textsuperscript{3$^\dagger$},
Zhanqiu Zhang\textsuperscript{3},\\
Yiwen Guo\textsuperscript{3$^\dagger$},
Chun Yuan\textsuperscript{2}, 
Ruifeng Xu\textsuperscript{1\thanks{\quad Corresponding authors.}}\\
\textsuperscript{1}Guangdong Provincial Key Laboratory of Novel Security Intelligence Technologies\\
\textsuperscript{2}Tsinghua Shenzhen International Graduate School\ \ \ \textsuperscript{3}Independent Researcher \\
\tt\small\{faris.xiong, rickywchen, zqzhang27, guoyiwen89, xuruifeng.hitsz\}@gmail.com \\
\tt\small crx23@mails.tsinghua.edu.cn; yuanc@sz.tsinghua.edu.cn\\
}

\begin{document}
\maketitle
\begin{abstract}
Model merging has recently gained attention as an economical and scalable approach to incorporate task-specific weights from various tasks into a unified multi-task model.
For example, in Task Arithmetic (TA), adding the fine-tuned weights of different tasks can enhance the model's performance on those tasks, while subtracting them leads to task forgetting. Although TA is highly effective, interference among task still hampers the performance of the merged model. Existing methods for handling conflicts between task generally rely on empirical selection, resulting in suboptimal performance.
In this paper, we introduce an Adaptive Weight Disentanglement method. We begin by theoretically proving that task vectors employed in model merging should be orthogonal to minimize interference among tasks. Guided by this insight, we initialize redundant vectors such that, when subtracted from the original task vectors, the resulting vectors exhibit increased orthogonality. Additionally, we impose an norm constraint on the redundant vectors to preserve the performance of the task-specific models. Experimental results demonstrate the effectiveness of our proposed technique: it successfully extracts redundant vectors, and after their subtraction, the task vectors not only retain robust performance but also achieve superior fusion outcomes.
Our code is available at \href{https://github.com/FarisXiong/AWD.git}{https://github.com/FarisXiong/AWD.git}.

\end{abstract}
    
\section{Introduction}
\label{sec:intro}

As the pretraining-finetuning paradigm gains increasing popularity~\cite{Active_Finetuning}, the research community has witnessed a proliferation of finetuned models~\cite{muqeeth2024learning}, typically derived from foundational models such as T5~\cite{t5}, and CLIP~\cite{clip}, among others. However, these models are often finetuned on task-specific training data, which limits their capacity for out-of-domain generalization~\cite{sanh2022multitask, uppaal-etal-2023-fine, UL2}. 
In diverse real-world applications, the independent deployment of multiple fine-tuned models increases storage costs and computational demands.
While traditional multi-task learning methods can mitigate these issues, they typically necessitate concurrent training across multiple task-specific datasets. 
Nonetheless, access to the original datasets is frequently restricted, and managing such datasets incurs significant expenses and potential privacy risks~\cite{mtl_protected}. 
Moreover, when confronted with new tasks, traditional multi-task learning methods necessitate training from scratch. Consequently, exploiting existing models to construct efficient multi-task models has become a crucial challenge.
While traditional multi-task learning methods can mitigate these issues, they typically require simultaneous training on multiple task-specific datasets, which poses several challenges. Access to these datasets is often restricted, and even when available, managing them incurs substantial costs and potential privacy risks~\cite{mtl_protected, yang2024modelmergingllmsmllms}.

Fortunately, model merging~\cite{task_arithmetic} has garnered growing attention as an economical and efficient method for obtaining multi-task models. 
This approach aims to merge multiple task-specific models, requiring no original training data or only a small amount of unlabeled data, thereby enabling the merged model to perform efficiently across diverse tasks~\cite{yang2024modelmergingllmsmllms}. 
As a foundational technique in this field,~\citet{task_arithmetic} introduced the concept of Task Arithmetic. Specifically, Task Arithmetic combines task vectors through arithmetic operations, facilitating the efficient transfer of capabilities among different tasks and thus enabling the construction of multi-task models.
Here, task vectors are derived by computing the difference between the weights of fine-tuned models and the pre-trained model.
However, related studies have indicated that task interference has become the main challenge for this method~\cite{ta_ntk,ties_merging}. 
Previous studies have proposed various techniques to mitigate this issue~\cite{ties_merging, dare, tall_mask, localizeandstitch}. 
Ties-Merging~\cite{ties_merging} addresses this challenge by pruning redundant parameters, resolving sign conflicts, and averaging parameters aligned along dominant directions. DARE~\cite{dare} reduces merging conflicts through random parameter drop and maintains model performance via rescaling operations. 
Consensus Merging~\cite{tall_mask} aims to eliminate selfish and catastrophic weights. 
Despite these advancements, these methods often rely on empirical strategies to resolve conflicts, lacking explicit optimization objectives, which leads to less than satisfactory performance.

In this paper, we revisit the Task Arithmetic Property and find that solving for task vectors satisfying this property in post-hoc methods is challenging. 
To more effectively guide task vector refinement, we propose the Task Consistency Property from the perspective of the merged model's performance. 
Specifically, this property requires that the performance of the merged model on each task should be close to that achieved when only the corresponding task vector is incorporated.
In this view, networks that satisfy this property can be considered as being free from interference between tasks.
Through theoretical derivation, we show that task consistency property can be approximately achieved by seeking orthogonal task vectors.
This solution is also partially supported by the experimental phenomenon of prior studies~\cite{task_arithmetic,l-lora}, which indicate that a smaller cosine similarity between task vectors leads to reduced interference among them, thereby enhancing the performance of the merged model.
Based on the previous discussions, we propose \textbf{A}daptive \textbf{W}eight \textbf{D}isentanglement (AWD). 
AWD aims to decompose traditional task vectors into a redundant vector and several disentangled task vectors, ensuring the disentangled task vectors can (1) exhibit enhanced orthogonality mutually while (2) maintaining the performance of specific tasks.
It achieves these two characteristics by two key optimization objectives respectively: (1) minimizing the cosine similarity of the disentangled task vectors; and (2) minimizing the norm of the redundant vector.
Furthermore, AWD can be seamlessly integrated into existing merging methods, such as Task Arithmetic~\cite{task_arithmetic} and AdaMerging~\cite{adamerging}.

Extensive experimental validations have demonstrated AWD’s effectiveness, and outperform existing merging techniques under various circumstances. 
When integrated with TA, AWD achieved absolute improvements in average accuracy of 2.8\% on the ViT-B/32 model and 1.5\% on the ViT-L/14 model compared to the advanced merging methods Ties-Merging.
Moreover, we conducted further analyses, demonstrating that our method generalizes effectively on language models and exhibits enhanced robustness under various conditions.
Additionally, we visualized the loss landscapes of joint tasks, which strongly supports our theoretical results.
In summary, the main contributions of this research can be summarized as follows:

\begin{itemize}
    \item We revisit Task Arithmetic Property and introduce appropriate relaxations by proposing Task Consistency Property. Based on this, we consider fixing pre-trained model and solving for task vectors that satisfy this property. Furthermore, we derive that the task vectors corresponding to different tasks should be mutually orthogonal.
    \item We propose a novel Adaptive Weight Disentanglement method that identifies and removes redundant vectors through the joint optimization of the orthogonality among disentangled task vectors and the norm of the redundant vectors. By subtracting these redundant vectors from each task vector, 
    our disentangled task vectors significantly reduce interference between tasks during model merging while preserving their task-specific performance.
    \item We conducted extensive evaluations on multiple benchmarks in computer vision and natural language processing to validate the advantages of our method. By combining our approach with existing methods, our AWD consistently improves upon previous merging methods, achieving state-of-the-art results.
\end{itemize}

\section{Related Work}
\label{sec:related_work}

\noindent\textbf{Orthogonal Optimization in Continual Learning:} 
The primary challenge in continual learning is enabling models to acquire the capability of new tasks without forgetting previously learned knowledge~\cite{continual_learning_survey_2024}. Recent studies \citep{zeng2019continual,farajtabar2020orthogonal,wang2023orthogonal,AOP_2022,HBO_2024} have demonstrated that maintaining the orthogonality of gradients during parameter updates can effectively mitigate interference with previous tasks in continual learning.
\citet{zeng2019continual} introduced an optimization technique based on orthogonal projection, which effectively reduces forgetting by projecting the gradient of new tasks onto the orthogonal subspace of the gradient of previous tasks during parameter updates. 
Similarly, \citet{farajtabar2020orthogonal} proposed Orthogonal Gradient Descent with Memory, enhancing the continual learning capability of models by integrating a memory replay mechanism. 
\citet{SGP_2023} introduce a scaled gradient projection method that integrates orthogonal projection with scaled updates along significant past gradients, thereby achieving enhanced task generalization with minimal forgetting.
\citet{wang2023orthogonal} propose optimizing LoRA within the orthogonal subspace to preserve the generalization capabilities.
Orthogonal optimization methods have demonstrated unique advantages for avoiding parameter and gradient conflict in continual learning. From this perspective, orthogonal task vectors can be further considered as a potential solution to model merging, as exemplified by our AWD approach.

\noindent\textbf{Multi-task Model Merging:} 
Model merging encompasses a diverse set of methodologies, including weight alignment~\cite{git-Re-Basin, jordan2023repair}, architectural transformation~\cite{fusellm}, multi-task learning~\cite{task_arithmetic, ta_ntk}, knowledge editing~\cite{language_ta_safety}, and other related techniques~\cite{yang2024modelmergingllmsmllms}.
In this paper, we concentrate on the multi-task model merging, which can be categorized into pre-hoc and post-hoc approaches.
\textbf{For pre-hoc methods:} 
Pre-hoc methods advocate modifying the model’s training procedure prior to training to enhance weight disentanglement. \citet{ta_ntk} established a connection between Task Arithmetic~\cite{task_arithmetic} and the spatial localization of NTK eigenfunctions, providing a method through NTK linearization to amplify weight disentanglement.
Building on this insight, ~\citet{l_lora} proposed L-LoRA, that partially linearizes adapter modules and applies arithmetic operations to the linearized adapters. This approach leverages the advantages of linearized fine-tuning for model merging while efficiently performing both fine-tuning and inference.
\textbf{For post-hoc methods:}
Early merging methods primarily focused on integrating individual models.
Simple Averaging~\cite{modelsoups} constructs the merged weights by independently computing the arithmetic mean of each corresponding parameter across all models.
Fisher Merging~\cite{fisher_merging} performs weighted parameter fusion by utilizing the fisher information matrix to assess the importance of individually fine-tuned model parameters. RegMean~\cite{regmean} addresses model merging by minimizing the predictive discrepancies between the merged model and the task-specific models.
Recently, \citet{task_arithmetic} demonstrated that efficient capability transfer can be achieved by combining task vectors through arithmetic operations.
However, task interference remains a significant challenge.
Ties-Merging~\cite{ties_merging} resolves this challenge by trimming redundant parameters, resolving the sign conflicts, and averaging parameters that align with the predominant direction.
DARE~\cite{dare} mitigates merging conflicts by randomly dropping parameters and preserves model performance through essential unscaling operations.
Consensus Merging~\cite{tall_mask} eliminates selfish and catastrophic weights, thereby enhancing the overall performance of existing model merging methods while simultaneously compressing the model.
In our work, we have demonstrated that the orthogonality among task vectors is the key to improving performance in model merging and introducing adaptive weight disentanglement to improve orthogonality.

\section{Methodology}

\label{sec:Method}

\subsection{Preliminary of Multi-task Model Merging}

\noindent\textbf{Notations:}  
Formally, we define the weights of pre-trained model by $\Theta$, which is initially trained on large-scale datasets to acquire general capabilities. 
Consider $K$ tasks, for each task, the pre-trained parameters $\Theta$ are fine-tuned using a domain-specific training dataset, resulting in task-specific model sets denoted as $\mathcal{M} = \{\Theta_i^\star\}_{i=1}^{K}$.

\noindent\textbf{Task Vectors:}  
Building on this,~\citet{task_arithmetic} define the task vector $\tau_i = \Theta_i^\star - \Theta$, which represents the task-specific adaptations from the fine-tuning process. We define the task vectors set as $\mathcal{T} = \{\tau_i\}_{i=1}^{K}$.

\noindent\textbf{Multi-task Model Merging:}  
The objective of multi-task model merging is to combine the task-specific models set $\{\Theta_{i}^\star\}_{i=1}^{K}$ into a unified model $\widehat{\Theta}$, or to merge the task vectors $\mathcal{T}$ into the pre-trained model $\Theta$. The merged model $\widehat{\Theta}$ aims to generalize effectively across all $K$ tasks, without resorting to approaches such as retraining from scratch or requiring full access to the training datasets of all tasks. 
Following \citet{task_arithmetic, fisher_merging, ties_merging}, we focus on merging task vectors into the pre-trained model.

\noindent\textbf{Task Arithmetic Property:} 
In context of multi-task model merging, we aspire adding task vector will not affect the output in other domains, which is defined as Task Arithmetic Property by~\citet{ta_ntk}. 

\noindent\textbf{Property 1} (Task Arithmetic Property). \emph{Given coefficient sets $\{\lambda_i\}_{i=1}^K\subset \mathbb{R}$ and a set of task vectors $\mathcal{T} = \{\tau_i\}_{i=1}^{K}$ correspond with non-intersecting task-specific data supports $\mathcal{D}=\{\mathcal{D}_i\}_{i \in [K]}$, i.e., $\forall t,t^\prime$, if $t\neq t^\prime$ then $\mathcal{D}_t\cap\mathcal{D}_{t^\prime}=\varnothing$. We say a
 network $f$ satisfies the task arithmetic property around $\Theta$ with respect to $\mathcal{T}$ and $\mathcal{D}$ if:}
\begin{align}
    f\left(x;\Theta+\sum_{i}^{K}\lambda_i\tau_i\right)=\begin{cases}
    f(x;\Theta+\lambda_i\tau_i) & x\in \mathcal{D}_{i} \\
    f(x;\Theta) & x\notin\bigcup_{i=1}^{K} \mathcal{D}_{i} \end{cases}.
    \label{eq:task_arithmetic}
\end{align}

\subsection{Theoretical Analysis}
\label{theory}

In an ideal scenario, we desire the network to possess task arithmetic property~\ref{eq:task_arithmetic}. 
Given that the output of a model typically depends on its input, it is challenging to retrospectively adjust the pre-trained model or task vectors to satisfies this property.
Considering the fundamental requirement for model merging is that the merged model should demonstrate performance on each task comparable to that of its respective task-specific model, we propose the \textbf{Task Consistency Property}.
Specifically, this property necessitates that the performance of the merged model on each task should be close to the performance achieved when only the corresponding task vector is integrated.
In this context, the networks that satisfy task consistency property can be considered free from interference between tasks. 
Task consistency property can be regarded as a more relaxed extension of the task arithmetic property, but it aligns better with the goal of multi-task model merging and is easier to solve. Its formal expression is as follows:

\noindent\textbf{Property 2} (Task Consistency Property). \emph{Given coefficient sets $\{\lambda_i\}_{i=1}^K\subset \mathbb{R}$ and a set of task vectors $\mathcal{T} = \{\tau_i\}_{i=1}^{K}$ correspond with non-intersecting task-specific data supports $\mathcal{D}=\{\mathcal{D}_i\}_{i \in [K]}$, i.e., $\forall t,t^\prime$, if $t\neq t^\prime$ then $\mathcal{D}_t\cap\mathcal{D}_{t^\prime}=\varnothing$. We say a network $f$ satisfies the task consistency property around $\Theta$ with respect to $\mathcal{T}$ and $\mathcal{D}$ if:}
\begin{align} 
\forall \; \mathcal{D}_i\in \mathcal{D} \; , \; \mathcal{L}_i\left(\Theta+\sum_{j}^{K}\lambda_j\tau_j\right)=
    \mathcal{L}_i\left(\Theta+\lambda_i\tau_i\right),
    \label{eq:task_consistency}
\end{align}
where $\mathcal{L}_i(\cdot)$ denotes the loss function of task $i$. 
Property~\ref{eq:task_arithmetic} and~\ref{eq:task_consistency} are characteristics jointly manifested by the pre-trained model and the task vectors. 
Considering that the pre-trained model serves as the cornerstone of model merging~\cite{ta_ntk}, simultaneously editing both the pre-trained model and the task vectors would hinder the network's ability to scale to new tasks. 
Therefore, we fix the pre-trained model and solve for the task vector that satisfies Property~\ref{eq:task_consistency}, which guides us in refining the task vector.

\begin{figure*}[!ht]
\centering
\includegraphics[width=0.9\linewidth]{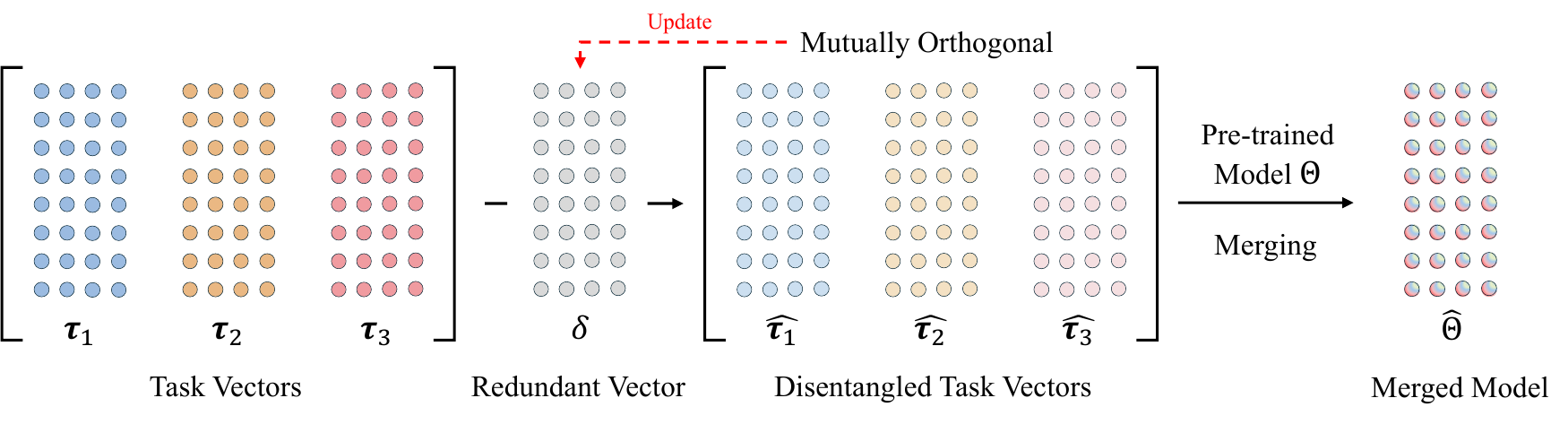}
\caption{Illustration of our Adaptive Weight Disentanglement.}
\label{method}
\end{figure*}

To simplify Eq.~\ref{eq:task_consistency}, we formally define the Merging Gap $G_i$ for task $i$ as:
\begin{align}
    G_{i} = \mathcal{L}_i \left( \Theta  + \sum_{j}^{K}\lambda_j\tau_j \right)  - \mathcal{L}_i \left( \Theta + \lambda_i\tau_i \right).
    \label{merging_error}
\end{align}

Based on this, the task consistency property can be rewritten as follows,
\begin{align}
    \forall \;i,\; G_{i} = 0.
    \label{eq:rewritten}
\end{align}

We then apply Taylor expansion around $\Theta$ on the right-hand side of Eq.~\ref{merging_error}. 
Given that the task vectors typically have small absolute values~\cite{dare}, terms of second-order and higher are generally negligible.
Therefore, we use a first-order approximation and simplify the expression, which could be described as follows, 
\begin{align}
G_i & \approx \left( \mathcal{L}_i \left( \Theta  \right) + \left< \nabla_{\Theta} \mathcal{L}_i(\Theta), \sum_{j}^n \lambda_j\tau_j \right> \right) \\
& \qquad -  \left( \mathcal{L}_i \left( \Theta  \right)
+ \left< \nabla_{\Theta} \mathcal{L}_i(\Theta), \lambda_i\tau_i  \right> \right) \\
& = \sum_{j\neq i}^{K}\left< \nabla_{\Theta} \mathcal{L}_i(\Theta), \lambda_j\tau_j \right>.
\end{align}

Since the original data is often difficult to obtain or may not be accessible, calculating the gradient of the pre-trained model for a single task is challenging. As an alternative, we can replace this gradient with the task vector, as the task vector can be interpreted as the accumulation of gradients. Therefore, the gradient $\nabla_{\Theta} \mathcal{L}_i(\Theta)$ can be estimated as $k_i  \tau_i$, where $k_i < 0$. Therefore, the merging gap $G_i$ can be estimated as:
\begin{align}
   G_i \approx \sum_{j\neq i}^{K}\left< k_i\tau_i,  \lambda_j\tau_j\right>,\\
   = k_i\sum_{j\neq i}^{K}\lambda_j\left< \tau_i, \tau_j\right>.
   \label{final_obj}
\end{align}

Due to the inability to compute the magnitudes of the gradients, solving the aforementioned equation becomes challenging. As a result, we aim to find a particular solution for the equation. As shown in Eq.~\ref{final_obj}, when all task vectors are orthogonal to each other, Eq.~\ref{eq:rewritten} will be satisfied: 
\begin{align}
   \forall \; i\ne j, \; \cos \left ( \tau_i,\tau_j  \right ) = 0  \Rightarrow \forall i, \; G_i = 0.
\end{align}
Our derivation results are partially supported by the experimental phenomenon of previous studies~\cite{task_arithmetic,l-lora}, which indicates that smaller positive cosine similarity leads to better performance in multi-task merging.
Consequently, based on the aforementioned discussions, our work endeavors to identify a novel task vector set of $\hat{\mathcal{T}}=\{\hat{\tau}_{1},\hat{\tau}_{2},...,\hat{\tau}_{K}\}$, such that $\Theta+\hat{\tau}_{i}$ optimally restores the performance of the fine-tuned model $\Theta_{i}^\star$, while simultaneously enhancing the orthogonality between task vectors $\hat{\tau}_i$ and $\hat{\tau}_j$.

\subsection{Adaptive Weight Disentanglement}

Previous studies have shown that task vectors exhibit significant redundancy~\cite{dare, ties_merging}. By eliminating redundant components, some conflicts along the parameter directions can be avoided, which helps reduce interference between tasks. 
Inspired by this, our proposed AWD approach aims to identify and remove redundant vector from a set of task vectors to obtain more effective task vectors.
As illustrated in Figure \ref{method}, when decomposing task vectors into redundant vector and disentangled task vectors, we expect the following characteristics could be satisfied. (1) \textbf{Orthogonality}: the disentangled task vectors exhibit smaller cosine similarity; (2) \textbf{Invariance}: the performance of the disentangled task vectors on their respective tasks remains comparable to that of the original task vectors. In this section, we define these two characteristics as our optimization objectives and utilize automatic differentiation tools to solve the aforementioned objectives \textbf{Adaptively}.

Formally, we define the redundant vector as $\delta$, which is a trainable vector variable initialized to zero. By subtracting $\delta$ from each task vector, we obtain the disentangled task vectors set $\widehat{\mathcal{T}}=\{\widehat{\tau_i} \}_{i=1}^{K}$, where $\widehat{\tau_i}=\tau_i-\delta$. Based on this, the first optimization objective for \textbf{Orthogonality} can be expressed as:
\begin{align}
\mathcal{L}_{\mathcal{O}} &= \frac{1}{K \left(K-1\right)}\sum_{i}^{K} \sum_{j \neq i}^{K} \left| \mathcal{F}\left( \widehat{\tau_i},\ \widehat{\tau_j} \right) \right|    \\
&= \frac{1}{K \left(K -1\right)}\sum_{i}^{K} \sum_{j \neq i}^{K} \left| \mathcal{F}\left( \tau_i-\delta,\ \tau_j-\delta \right) \right|,
\label{objective_orthogonality}
\end{align}
where $\mathcal{F}$ denotes the cosines similarity function.

Besides, without proper constraints on $\delta$, the disentangled task vector $\tau_i - \delta$ may experience a significant drop in performance on their corresponding tasks. This is because an excessively large redundant vector may remove information within the task vectors that is crucial for task performance.
Recent studies have demonstrated that pruning the smaller components of the task vector has a negligible impact on model performance~\cite{ties_merging, dare}. 
Motivated by this insight, we introduce an norm constraint on $\delta$ and incorporate it as the second optimization objective for \textbf{Invariance}, which is formally expressed as: \footnote{Please refer to the Section~\ref{invariance_derivation} for a detailed proof.}
\begin{align}
\mathcal{L}_{\mathcal{R}} = \left|\left| \delta \right|\right|.
\label{invariance_orthogonality}
\end{align}

Considering the above, our final optimization objective function can be expressed as a weighted sum of these two parts, formally,
\begin{align}
\mathcal{L}  = \mathcal{L}_{\mathcal{O}} + \alpha \mathcal{L}_{\mathcal{R}},
\label{L_obejective}
\end{align}
where $\alpha$ is a hyperparameter used to balance the relationship between orthogonality and invariance. By minimizing the total loss, we can obtain the redundant vector $\delta = \operatorname*{arg\,min}_{\delta} \mathcal{L}\left(\delta; \mathcal{T}\right)$.
Then we can get the disentangled task vector by subtracting the redundant vector from the task vector $\widehat{\tau_i} = \tau_i - \delta$.

Our approach can be seamlessly integrated into existing model merging approaches, including Task Arithmetic~\cite{task_arithmetic} and AdaMerging~\cite{adamerging}. 
For \textbf{Task Arithmetic:} the final merged model $\widehat{\Theta}$ can be fomulated as $\widehat{\Theta} = \Theta + \lambda \sum_{i}^{K}\widehat{\tau_i}$.
For \textbf{AdaMerging:} the final merged model $\widehat{\Theta}$ can be fomulated as $\widehat{\Theta} = \Theta + \sum_{i}^{K}\sum_j^{\mathcal{P}}\lambda_{ij}\widehat{\tau_i}^{j}$, where $\mathcal{P}$ denotes the number of layers in $\widehat{\tau_i}$, and $\widehat{\tau_i}^j$ represents the parameter in $j$-th layer of $\widehat{\tau_i}$.

\begin{algorithm}[htbp]
\small
\caption{\small Adaptive Weight Disentanglement} %
\label{algorithm}
\KwIn{Task Vectors $\mathcal{T} = \{\tau_i\}_{i=1}^{K}$; Solution Steps $\mathcal{N}$; Learning Rate $\beta$; Hyperparameter $\alpha$.\\} %
\KwOut{Disentangled Task Vectors $\widehat{\mathcal{T}}$.} %
\LinesNumbered %
$\triangleright$ Initialize Redundant Vector $\delta^0$. \\
$\delta^0\leftarrow 0$

\For{$n\in\{1,\cdots,\mathcal{N}\}$}{
$\triangleright$ Initialize Total Loss $\mathcal{L}$. \\
$\mathcal{L}\leftarrow 0$

\For{$i\in\{1,\cdots,K\}$}{
\For{$j\in\{1,\cdots,K\}$}{
$\triangleright$ Calculate Loss by Redundant Vector and Task Vectors.\\
$\mathcal{L} \mathrel{+}= \mathcal{L}_\mathcal{O}\left(\delta^{n-1} ; \tau_i,\tau_j \right)$\\    
}
}
$\mathcal{L} \mathrel{+}= \alpha\mathcal{L}_\mathcal{R}$\\  
$\triangleright$ {Update} the Redundant Vector $\delta^n$. \\
$\delta^{n} = \delta^{n-1} - \beta\nabla_{\delta^{n-1}}\mathcal{L}\left(\delta^{n-1} ; \mathcal{T} \right)$\\
    
}
$\triangleright$ Calculate the Disentangled Task Vector.\\
$\widehat{\mathcal{T}} =  \{\tau_i - \delta^\mathcal{N}\}_{i=1}^{K}$ \\
\end{algorithm}

\section{Experiments}
In this section, we will detail the experimental setup and the analysis of the results of this study. More detailed experimental data and additional analyses will be included in the Supplementary Materials.

\begin{table*}[ht]
\centering
\caption{Multi-task performance when merging ViT-B/32 models on 8-task vision benchmark. The best performance across different merging methods is denoted in bold.}
\label{tab:performance_vitbase32} 
\resizebox{0.95\linewidth}{!}{  
\begin{tabular}{l|cccccccc|cc}
\toprule
\textbf{Method}  &  \textbf{SUN397}  &  \textbf{Cars}  &  \textbf{RESISC45}  &  \textbf{EuroSAT}  &  \textbf{SVHN}  &  \textbf{GTSRB}  &  \textbf{MNIST}  &  \textbf{DTD}  & \textbf{Avg Acc}  \\
\midrule
\multicolumn{10}{c}{\emph{Non-merging Methods}} \\
{Pretrained}  &  {62.3}  &  {59.7}  &  {60.7}  &  {45.5}  &  {31.4}  &  {32.6}  &  {48.5}  &  {43.8} & {48.0}   \\
{Individual}  &  79.2  &  77.7  &  96.1  &  99.7  &  97.5  &  98.7  &  99.7  &  79.4 & 90.8   \\
{Traditional MTL}    &  73.9  &  74.4  &  93.9  & 98.2    &  95.8  &  98.9   &  99.5   & 77.9 & 88.9  \\
\midrule
\multicolumn{10}{c}{\emph{Training-free Methods}} \\
{Weight Averaging} & 65.3  &  63.4  &  71.4  &  71.7  &  64.2  &  52.8  &  87.5  &  50.1  & 65.8 \\
{Fisher Merging} &  \textbf{68.6}  &  \textbf{69.2}  &  70.7  &  66.4  &  72.9  &  51.1  &  87.9  &  \textbf{59.9} & 68.3  \\
{RegMean}  &  65.3  &  63.5  &  {75.6}  &  78.6  &  78.1  &  67.4  &  93.7  &  52.0 & 71.8 \\
Task Arithmetic   &55.2 &54.9 &66.7 &78.9 &80.2 & 69.7 &97.3 &50.4 & 69.1 \\
{Ties-Merging}   &  59.8  &  58.6  &  70.7  &  79.7  &  \textbf{86.2}  &  72.1  &  \textbf{98.3}  &  54.2 & 72.4  \\
Consensus Merging & 65.7 & 63.6 & \textbf{76.5} & 77.2 & 81.7 & 70.3 & 97.0 & 57.1 & 73.6 \\
\textbf{AWD Task Arithmetic} (Ours) & 63.5 & 61.9  & 72.6 & \textbf{84.9} & 85.1 & \textbf{79.1} & 98.1 & 56.7 & \textbf{75.2}\\ %
\midrule
\multicolumn{10}{c}{\emph{Test-time Adaption Methods}} \\
{TW AdaMerging} &58.0 &53.2 &68.8 &85.7 &81.1 &84.4 &92.4  &44.8 &71.1 \\
{TW AdaMerging++} &60.8 &56.9 &73.1 &83.4 &87.3 &82.4 &95.7  &50.1 &73.7 \\
{LW AdaMerging}  &64.5 &68.1 &79.2 &93.8 &87.0 &91.9 &97.5  &59.1 &80.1 \\
{LW AdaMerging++} &66.6 &68.3 &82.2 &94.2 & \textbf{89.6} &89.0 &98.3  &60.6 &81.1 \\
{Representation Surgery}  & 63.8 & 59.9 & 83.3 & \textbf{97.9} & 87.0 & 87.0 & \textbf{98.6} &  \textbf{69.4} & 80.9 \\
\textbf{AWD AdaMerging} (Ours) & \textbf{68.1} & \textbf{71.4} & \textbf{83.4} & 94.8 & 87.7 & \textbf{93.6} & 97.9 & 66.1 & \textbf{82.9}\\ %
\bottomrule
\end{tabular}
}
\end{table*}

\begin{table*}[t]
\centering
\caption{Multi-task performance when merging ViT-L/14 models on 8-task vision benchmark. The best performance across different merging methods is denoted in bold.}
\label{tab:performance_vitlarge14} 
\resizebox{0.95\linewidth}{!}{  
\begin{tabular}{l|cccccccc|cc}
\toprule
\textbf{Method}  &  \textbf{SUN397}  &  \textbf{Cars}  &  \textbf{RESISC45}  &  \textbf{EuroSAT}  &  \textbf{SVHN}  &  \textbf{GTSRB}  &  \textbf{MNIST}  &  \textbf{DTD}  & \textbf{Avg Acc}  \\
\midrule
\multicolumn{10}{c}{\emph{Non-merging Methods}} \\
 {Pretrained}  & {66.8}  & {77.7}  & {71.0}  &   {59.9}  & {58.4}  &  {50.5}  & {76.3}  &   {55.3} &  {64.5}   
\\
{Individual}   &  82.3  &  92.4  &  97.4  &  100  &  98.1  &  99.2  &  99.7  &  84.1  & 94.2   \\
{Traditional MTL} &  80.8   &  90.6   &   96.3  & 96.3   & 97.6   & 99.1   &  99.6  &  84.4   & 93.5    \\
\midrule
\multicolumn{10}{c}{\emph{Training-free Methods}} \\
{Weight Averaging}    &  72.1  &  81.6  &  82.6  &  91.9  &  78.2  &  70.7  &  97.1  &  62.8 & 79.6 \\
{Fisher Merging}     &  69.2  &  \textbf{88.6}  &  87.5  &  93.5  &  80.6  &  74.8  &  93.3  &  \textbf{70.0}  & 82.2 \\
{RegMean}    &  73.3  &  81.8  &  86.1  &  \textbf{97.0}  &  88.0  &  84.2  &  98.5  &  60.8  & 83.7 \\
{Task Arithmetic}   &73.9  &82.1 &86.6 &94.1  &87.9  &86.7  &98.9  &65.6   &84.5 \\
{Ties-Merging}   &  \textbf{76.5}  &  85.0  &  {89.3}  &  95.7  &  90.3  &  83.3  &  99.0  &  68.8  & 86.0   \\
Consensus Merging & 75.0 & 84.3 & \textbf{89.4} & 95.6 & 88.3 & 82.4 & 98.9 & 68.0 & 85.2\\
\textbf{AWD Task Arithmetic} (ours) & 76.2 & 85.4 & 88.7 & 96.1 & \textbf{92.4} & \textbf{92.3} & \textbf{99.3} & 69.4 & \textbf{87.5}\\
\midrule
\multicolumn{10}{c}{\emph{Test-time Adaption Methods}} \\
AdaMerging &79.0 &90.3 &90.8 & 96.2 &93.4 &98.0  &99.0  &79.9  &90.8 \\
AdaMerging++  &79.4 &90.3 &91.6 &97.4 &93.4 &97.5 & 99.0 &79.2 & 91.0 \\
Representation Surgery & 75.7 &  84.4 &  \textbf{93.1} &  \textbf{98.8} &  91.3 &  93.4 &  99.1 &  76.1 &  89.0\\

\textbf{AWD AdaMerging} (Ours) & \textbf{79.8} & \textbf{90.6} & {91.8} & 97.0 & \textbf{93.9} & \textbf{98.4} & \textbf{99.2} & \textbf{81.1} & \textbf{91.5}\\ %
\bottomrule
\end{tabular}
}
\end{table*}

\begin{figure}[t]
\centering
\includegraphics[width=0.9\linewidth]{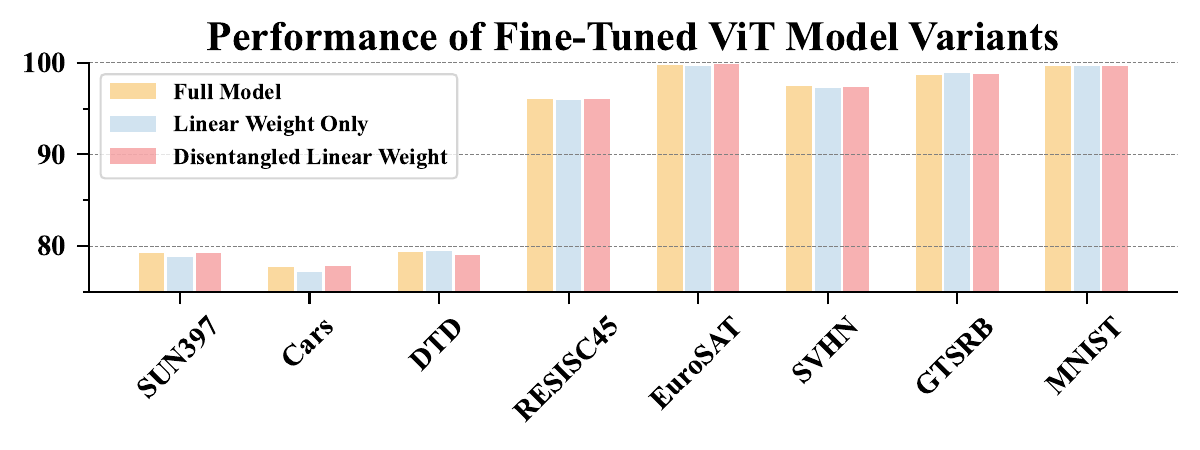}
\includegraphics[width=0.9\linewidth]{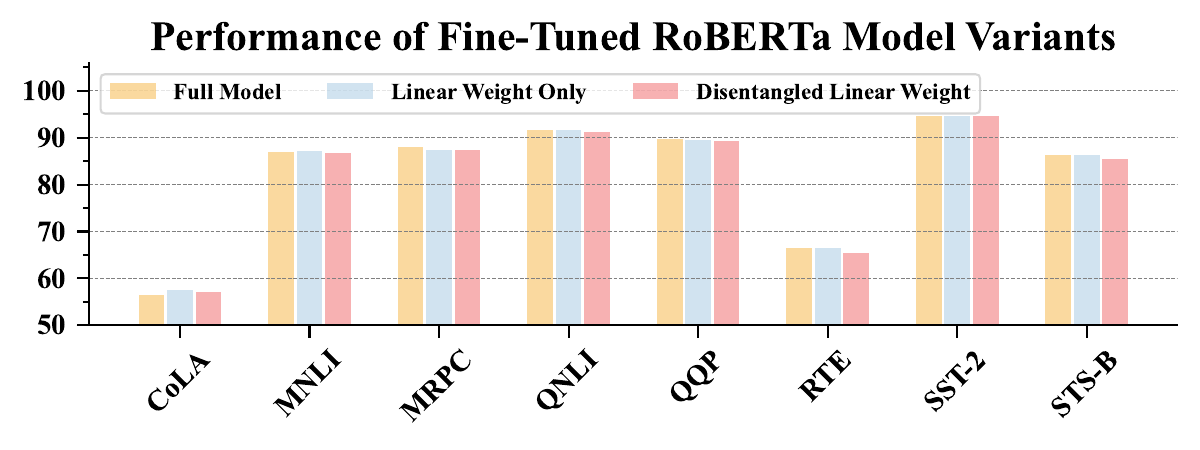}
\caption{Comparative Performance of Fine-Tuned ViT-B/32 and RoBERTa Model Variants.}
\label{pilot}
\end{figure}

\subsection{Experimental Settings}

\noindent\textbf{Datasets:} Following the previous studies~\cite{task_arithmetic,adamerging,ties_merging,surgery}, we explore multi-task model merging across eight image classification datasets: SUN397~\citep{xiao2016sun}, Cars~\citep{krause20133d}, RESISC45~\citep{cheng2017remote}, EuroSAT~\citep{helber2019eurosat}, SVHN~\citep{yuval2011reading}, GTSRB~\citep{stallkamp2011german}, MNIST~\citep{lecun1998mnist}, DTD~\citep{cimpoi2014describing}. 

\noindent\textbf{Models:}
For our experiments, we employ the ViT-B/32 and ViT-L/14 models, originally derived from CLIP~\cite{clip}. 

\noindent\textbf{Baselines:}
To ensure a comprehensive evaluation, we categorize the baselines into three main groups: Non-merging methods, Training-free methods, and Test-time Adaptation methods. The Non-merging category includes individual finetuned models and traditional multi-task learning approaches. The training-free methods we consider are Weight Averaging~\cite{modelsoups}, Fisher Merging~\cite{fisher_merging}, RegMean~\cite{regmean}, Task Arithmetic~\cite{task_arithmetic}, Ties-Merging~\cite{ties_merging}, and Consensus Merging~\cite{tall_mask}. Finally, we incorporate Test-time Adaptation methods, including AdaMerging~\cite{adamerging} and Representation Surgery~\cite{surgery}.

\begin{table}[ht]
\centering
\caption{Multi-task performance when merging RoBERTa models on 8-task GLUE benchmark. Following~\citet{Lu2024TwinMerging}, we report average normalized score. Red text indicates the performance improvements compared to the most advanced baseline. The best performance across different merging methods is denoted in bold.}
\label{tab:performance_roberta_all} 
\resizebox{\linewidth}{!}{  
\begin{tabular}{l|cc}
\toprule
\textbf{Method} & \textbf{RoBERTa-Base} & \textbf{RoBERTa-Large} \\

\midrule
Pretrained &  41.7 & 38.2\\
{Individual} & 100.0  & 100.0 \\
\midrule
Weight Averaging & 52.6 & 53.3\\
Task Arithmetic & 67.8 & 70.9\\
{Ties-Merging} & 64.7 & 72.4\\
Task Arithmetic (w/ DARE) & 63.7 & 70.9\\
{Ties-Merging} (w/ DARE) & 65.6 & 72.8\\
\textbf{AWD Task Arithmetic} (Ours)&  $\textbf{68.3}_{\textcolor{red}{\bigtriangleup 0.5}}$ & $\textbf{74.5}_{\textcolor{red}{\bigtriangleup 1.7}}$\\
\bottomrule
\end{tabular}
}
\end{table}

\subsection{Pilot Experiment}

Although our method is relatively simple, its applicability may still be limited when confronted with complex transformer-based models, particularly when generalizing our approach to language models with high-dimensional embedding layers~\cite{word_embedding_dimension}.
~\citet{dai-etal-2022-knowledge} suggest that knowledge neurons are stored within Feedforward Neural Networks.
Inspired by this, we have further simplified our approach: restricting the parameters of the task vector solely to the weights of the linear layers within the transformer blocks. 
To validate the rationality of this simplified strategy, we conducted a series of pilot experiments. 
As shown in Figure~\ref{pilot}, whether in vision models or language models, we found that preserving only the parameters of weights in the linear layer of the task vector did not affect the performance of the models. 
Moreover, even after applying our method to combine with the disentangled task vector in this simplified variant, the models were able to recover to a performance level similar to that of the full fine-tuned models.

\subsection{Main Results}

The performance of all baselines using the ViT-B/32 and ViT-L/14 architectures is presented in Table \ref{tab:performance_vitbase32} and Table \ref{tab:performance_vitlarge14}, respectively. 
Our observations are as follows:
(1) Our experiments demonstrate that individually fine-tuned models achieve the highest performance. Traditional multi-task learning approaches utilizing joint training exhibit slightly lower performance. 
(2) It is noteworthy that compared to previous merging methods, our strategy demonstrates significant performance improvements in consistency. By integrating Task Arithmetic, our approach achieves a notable improvement over the advanced method of Ties-Merging, outperforming it by 2.8\% on the ViT-B/32 model and by 1.5\% on the ViT-L/14 model. 
Moreover, in the test-time adaptation setting, by incorporating AdaMerging, our approach outperforms AdaMerging++ by 1.8\% on ViT-B/32 and by 0.5\% on ViT-L/14.
(3) As model sizes increase, the performance of merging methods progressively aligns with that of traditional multi-task learning methods. Our approach demonstrates consistent improvements. Specifically, on ViT-L/14, under the test-time adaptation setting, it achieves performance that is only 2\% lower than that of traditional multi-task learning.
(4) Compared to test-time adaptation, our method demonstrates a more significant performance improvement in training-free scenarios. We attribute this phenomenon to the dynamic adjustment of task coefficients within the test adaptation setting. Specifically, the coefficients of task vectors can be interpreted as indicators of each task vector's relative importance. Therefore, when the coefficients of task vectors change, it is appropriate to adjust the weight distribution in the optimization objective accordingly.

\subsection{Generalizability on Language Models}
We further extend our methodology to the language model RoBERTa~\cite{roberta} and utilize the GLUE benchmark~\cite{glue} to assess the generalizability of our approach.
Given the diverse evaluation metrics across tasks, we follow \citet{Lu2024TwinMerging} and report normalized score in Table \ref{tab:performance_roberta_all}, which leverages the fine-tuned models as upper bounds on performance to assess the effectiveness of the merging model.
Table \ref{tab:performance_roberta_all} presents the performance of various training-free methods for merging RoBERTa models across eight GLUE benchmark tasks. 
Specifically, our approach notably attains the highest average performance score of 68.3\% on RoBERTa-Base and 74.5\% on RoBERTa-Large, marking substantial enhancements over the baseline methods. Compared to the most advanced merging baseline, our methodology manifests a notable improvement of 0.5\% on RoBERTa-Base and a substantial 1.7\% increment on RoBERTa-Large.
These results highlight the superior generalization ability of our methods on language models.
Moreover, similar to the ViT series of vision models, we observe that the performance of the merge method approaches that of the individually fine-tuned models as the size of the model increases.

\begin{figure}[t]
\centering
\begin{subfigure}{0.48\linewidth}
    \includegraphics[width=\linewidth]{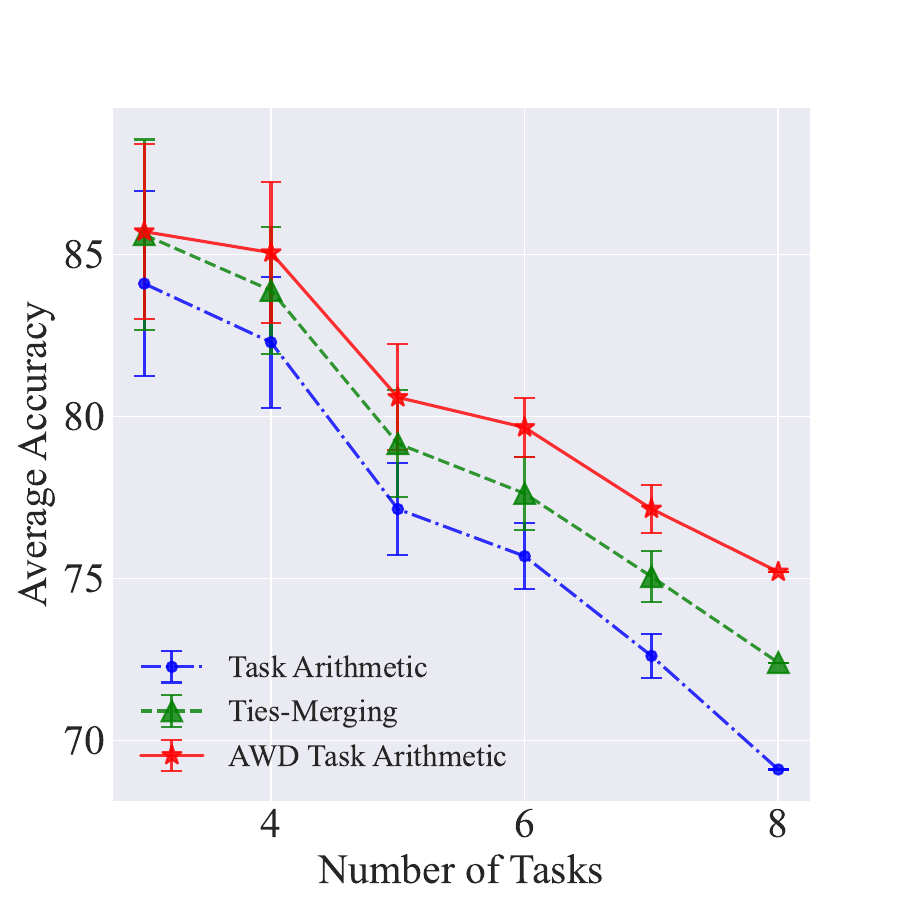}
    \subcaption{Average Accuracy Across Different Task Numbers}
    \label{robust_analysis_task_numbers}
\end{subfigure}
\hfill
\begin{subfigure}{0.48\linewidth}
    \includegraphics[width=\linewidth]{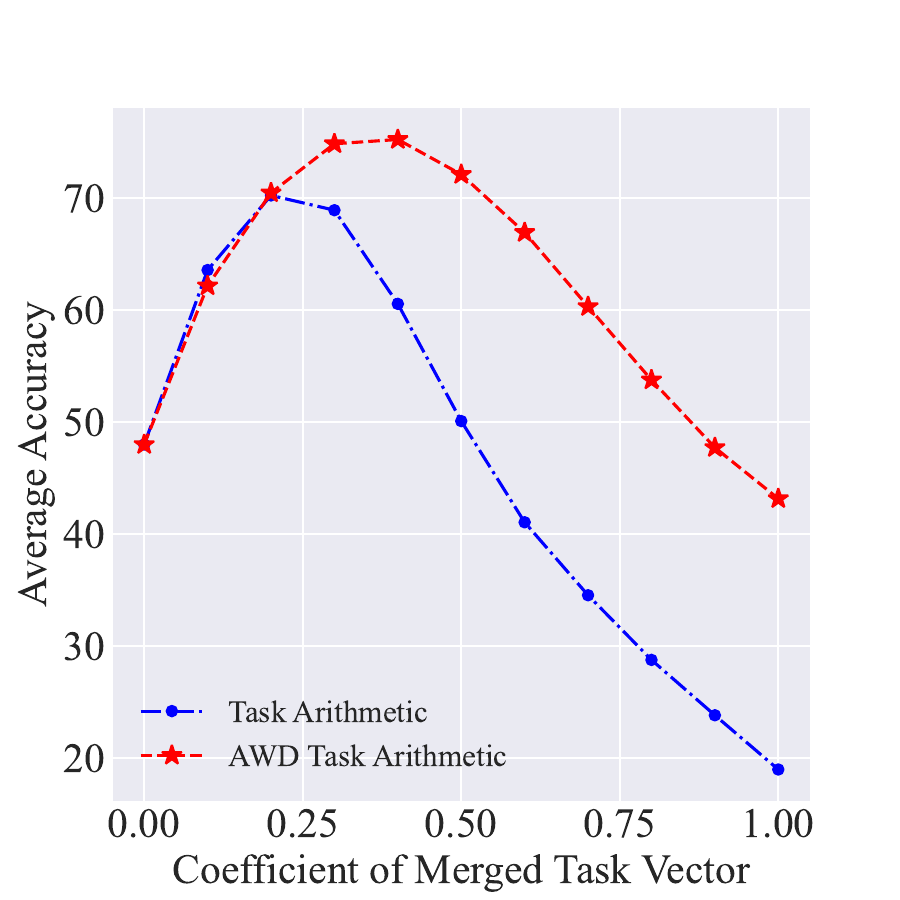}
    \subcaption{Average Accuracy Across Different Coefficients on 8 vision tasks}
    \label{robust_analysis_coefficients}
\end{subfigure}
\caption{Impact of task numbers and coefficients on average accuracy for ViT-B/32.}
\label{robust_analysis}
\end{figure}

\subsection{Robustness Analysis}
\label{robustness_analysis_text}
\noindent\textbf{Robustness Analysis on Task Numbers.}
We evaluated the performance of the merging model on in-domain tasks with varying numbers of tasks and compared the proposed method with Task Arithmetic and Ties-Merging.
Figure \ref{robust_analysis_task_numbers} shows the average accuracy of these methods across different numbers of merged tasks, with error bars representing the 95\% confidence interval.
For each setting with different task numbers, we randomly sampled 8 subsets from all tasks and computed the average accuracy of the merging model on these subsets. More experimental details can be found in the supplementary materials.
From Figure \ref{robust_analysis_task_numbers}, the following conclusions can be drawn: (1) As the number of merged tasks increases, the performance of all methods shows a declining trend, indicating that increased task numbers lead to greater interference between tasks; 
(2) As the number of merged tasks increases, our method's performance decreases more slowly compared to Task Arithmetic and Ties-Merging.
Overall, task interference remains the primary factor affecting model performance. 
Compared to the other baseline methods, our approach significantly reduces interference between task vectors, demonstrating superior robustness when handling a larger number of tasks.

\noindent\textbf{Robustness Analysis on Coefficients.}
In order to further assess the robustness of our method, we also conducted a comparative analysis of different coefficients for the merged task vectors. As shown in Figure \ref{robust_analysis_coefficients}, we compared our approach with Task Arithmetic. From the results presented in the figure, it can be observed that when the coefficient is less than or equal to 0.2, our method performs similarly to Task Arithmetic; however, when the coefficient exceeds 0.2, our method significantly outperforms Task Arithmetic. This indicates that as the task vectors are scaled up, the interference between tasks is also amplified. Nonetheless, our method can effectively eliminate inter-task interference, making it more robust to changes in the coefficients.

\begin{figure}[t]
\centering
\begin{subfigure}{1\linewidth}
    \includegraphics[width=\linewidth]{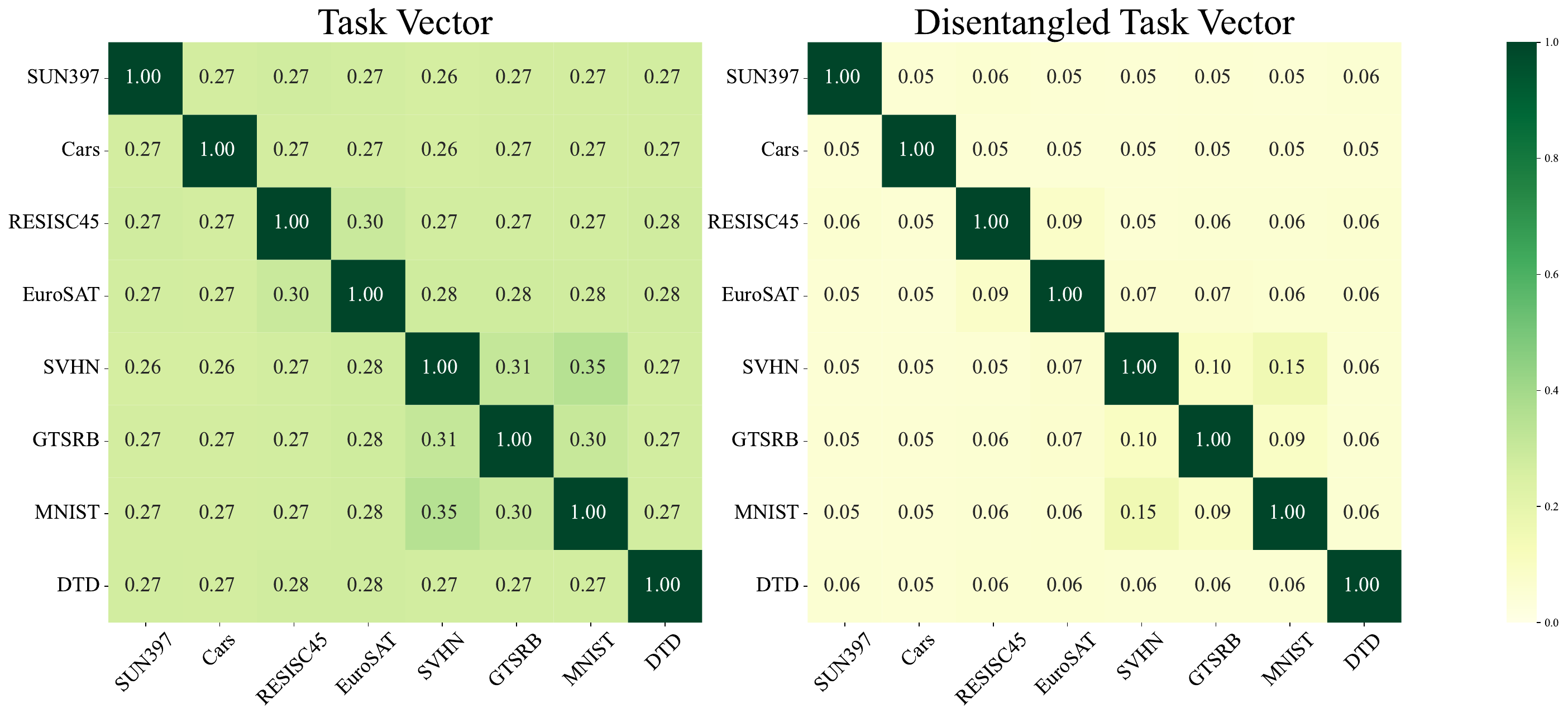}
    \subcaption{Cosine similarity between task vectors on ViT-B/32}
\end{subfigure}
\hfill
\begin{subfigure}{1\linewidth}
    \includegraphics[width=\linewidth]{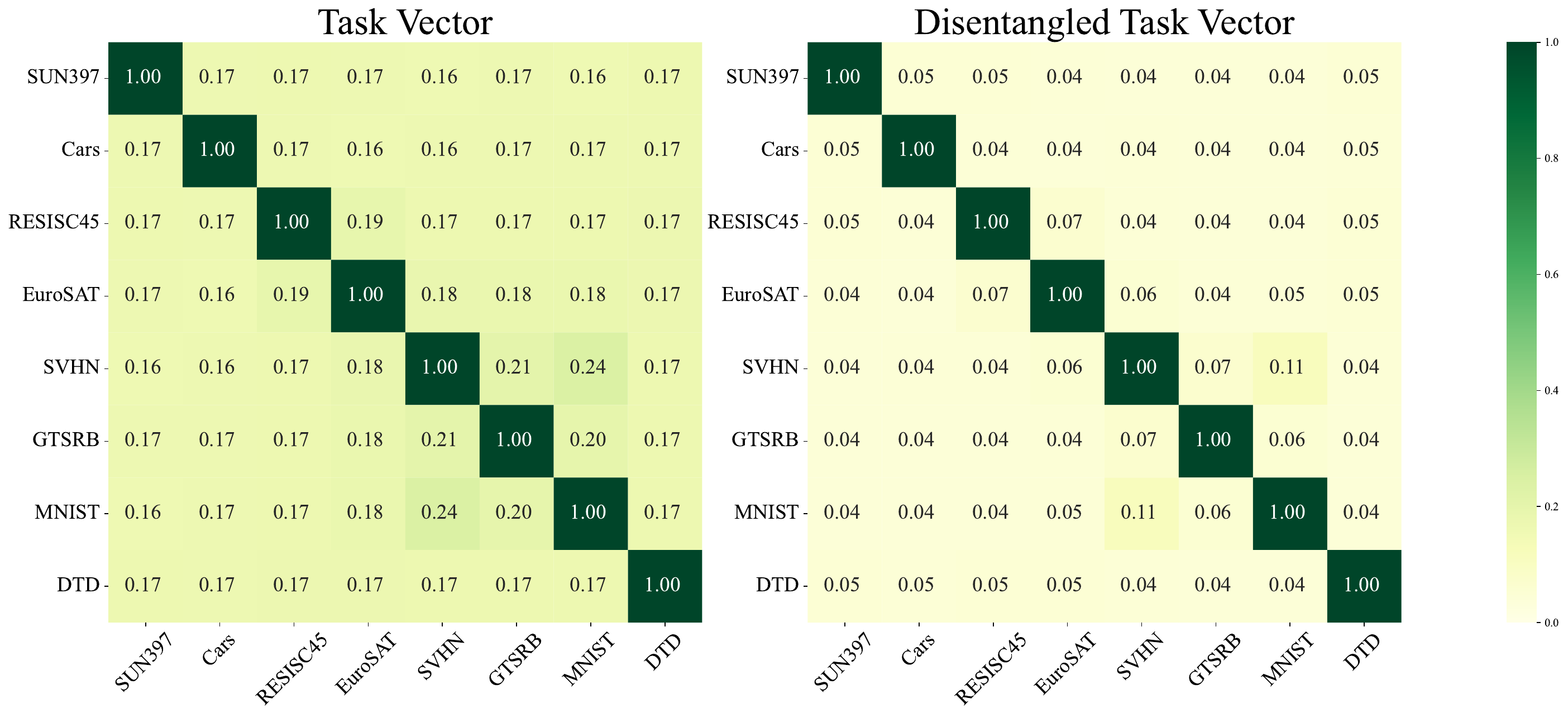}
    \subcaption{Cosine similarity between task vectors on ViT-L/14}
\end{subfigure}
\caption{Cosine similarity heatmaps for task vectors and disentangled task vectors on ViT-B/32 and ViT-L/14.}
\label{cosine_vit}
\end{figure}

\begin{figure}[t]
\centering
\begin{subfigure}{1\linewidth}
    \includegraphics[width=\linewidth]{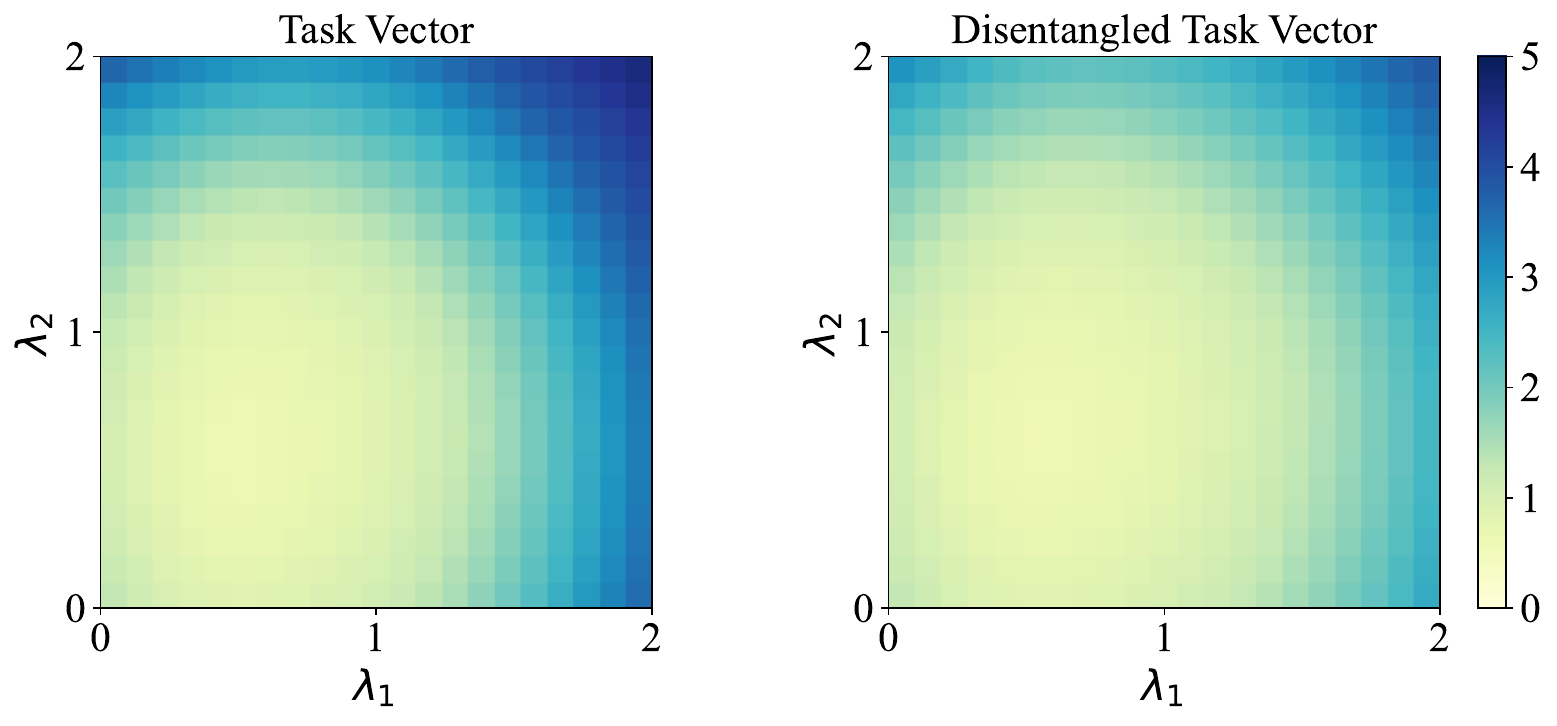}
    \subcaption{Loss landscape visualization on RESISC45 and Cars.}
    \label{loss_landscape_resisc45_cars}
\end{subfigure}
\hfill
\begin{subfigure}{1\linewidth}
    \includegraphics[width=\linewidth]{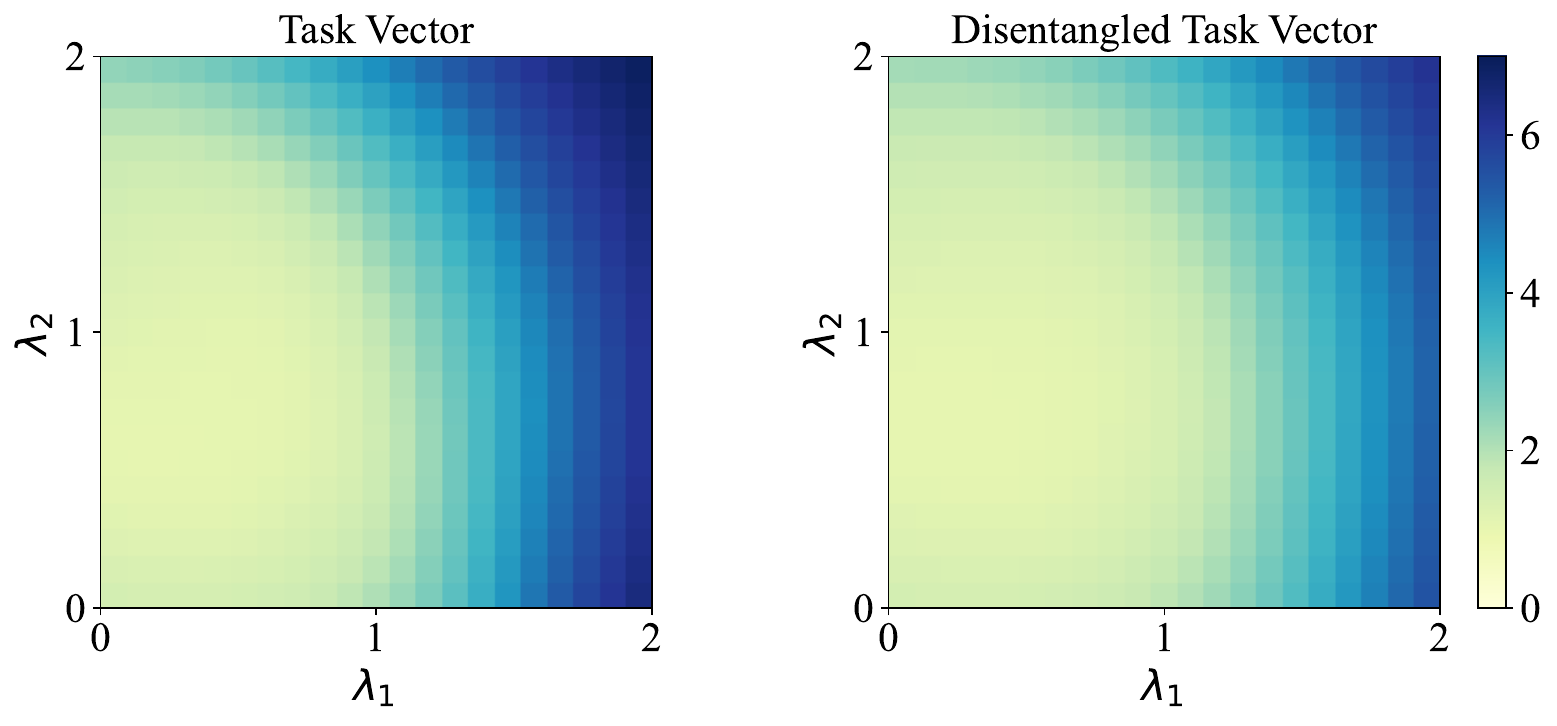}
    \subcaption{Loss landscape visualization on SUN397 and DTD.}
    \label{loss_landscape_sun397_dtd}
\end{subfigure}
\caption{Loss landscape visualization. We visualize the loss landscape $\mathcal{L}_i(\widehat{\Theta}) + \mathcal{L}_j(\widehat{\Theta}) $ by interpolating for ViT-B/32. %
}
\label{loss_landscape}
\end{figure}

\subsection{Visualization Analysis}

\noindent\textbf{Similarity between Task Vectors.}
As shown in Figure~\ref{cosine_vit}, we measure the cosine similarity of the task vectors for the ViT-B/32 and ViT-L/14 models, respectively. Compared with original task vectors, our method significantly reduces the cosine similarity between task vectors of different tasks, which is crucial for reducing interference between tasks~\cite{task_arithmetic, l_lora}. We find that, for larger models, the cosine similarity between task vectors is smaller. This may also be the key to the differences in task arithmetic performance among models of different sizes shown in Tables~\ref{tab:performance_vitbase32} and~\ref{tab:performance_vitlarge14}.

\noindent\textbf{Loss Landscape.}
To more clearly demonstrate the advantages of our AWD in reducing interference, we conducted a visualization analysis of the loss landscape for the joint tasks.
Specifically, we performed interpolations on a 2D plane between the pre-trained model $\Theta$ and the task vectors, and separately between the pre-trained model and the disentangled task vectors. 
As depicted in Figure~\ref{loss_landscape}, the heatmaps represent the loss values $\mathcal{L}_i(\widehat{\Theta})+\mathcal{L}_j(\widehat{\Theta})$ evaluated over the joint tasks. 
These figures show that the disentangled task vectors derived through our methodology exhibit a more expansive low-loss basin compared to the original task vector.
This demonstrates that our approach effectively reduces interference between tasks. Moreover, as $\lambda$ increases, regions initially shown in deep blue demonstrate a substantial reduction, indicating that our method is robust to variations in the coefficient $\lambda$, which is consistent with Section~\ref{robustness_analysis_text}.

\section{Conclusion}

In this paper, we propose the task consistency property, which requires that a merged model's performance on each task closely matches the performance achieved when only the corresponding task vector is incorporated.
In this context, task vectors satisfying this property are free from task interference. 
We theoretically demonstrate that this property can be approximately achieved by ensuring that task vectors are orthogonal. 
Based on this insight, we introduce Adaptive Weight Disentanglement, which decomposes traditional task vectors into a redundant vector and several disentangled task vectors, enhancing their orthogonality to approximate the solution of task consistency property.
Experimental results show that our approach effectively reduces interference between tasks and significantly improves the performance of existing merging techniques.

{
    \small
    \bibliographystyle{ieeenat_fullname}
    \bibliography{main}
}

\clearpage

\onecolumn

\maketitlesupplementary

\section{Further Theoretical Derivation}

\subsection{Analysis of Invariance Optimization Objective}\label{invariance_derivation}

Inspired by previous works~\cite{dare,ties_merging}, we empirically incorporate the norm of $\delta$ as \textbf{Invariance} objective to ensure that each disentangled task vector performs well on its corresponding specific task.
Herein, we further engage in a theoretical examination of the rationality of this optimization objective.
Ideally, we aim for each disentangled task vector to individually preserve the performance of the original task vector on its corresponding specific task.
We define the \textbf{Replacement Gap} $\mathcal{R}_i$ as the discrepancy in performance between the disentangled task vector $\widehat{\tau_i}$ and the original task vector $\tau_i$ on a specific task $i$. Then, $\mathcal{R}_i$ can be described as,
\begin{align}
    \mathcal{R}_i = \mathcal{L}_i\left(\Theta+\widehat{\tau_i} \right)-{\mathcal{L}_i} \left(\Theta+{\tau_i} \right).
    \label{R_i}
\end{align}

We apply the Taylor expansion to $\mathcal{L}_i(\Theta + \widehat{\tau_i})$ and $\mathcal{L}_i(\Theta + \tau_i)$. 
Taking $\mathcal{L}_i(\Theta + \widehat{\tau_i})$ as an example,
\begin{align}
    \mathcal{L}_i(\Theta + \widehat{\tau_i}) \approx \mathcal{L}_i(\Theta) + \left< \nabla_{\Theta} \mathcal{L}_i(\Theta), \widehat{\tau_i} \right> + \frac{1}{2}\widehat{\tau_i}^\top \mathbf{H}_i\widehat{\tau_i},
\end{align}
where $\mathbf{H}_i$ is the Hessian of the loss $\mathcal{L}_i$ at $\Theta$.
Given that $\tau_i$ is generally negligible~\cite{dare}, the second-order terms can typically be approximated as zero.
As a result, Equation~\ref{R_i} can be simplified accordingly.
\begin{align}
    \mathcal{R}_i &\approx \mathcal{L}_i(\Theta) + \left< \nabla_{\Theta} \mathcal{L}_i(\Theta), \widehat{\tau_i} \right> - \mathcal{L}_i(\Theta) - \left< \nabla_{\Theta} \mathcal{L}_i(\Theta), {\tau_i} \right> \\
    &= \left< \nabla_{\Theta} \mathcal{L}_i(\Theta), \widehat{\tau_i} - {\tau_i} \right> \\
    &= \left< \nabla_{\Theta} \mathcal{L}_i(\Theta), {\tau_i} - \delta - {\tau_i}\right> \\
    &= \left< \nabla_{\Theta} \mathcal{L}_i(\Theta), -\delta\right> \\
    &= \left|\nabla_\Theta \mathcal{L}_i(\Theta)\right|\left|-\delta\right|\cos\left(\nabla_\Theta\mathcal{L}_i(\Theta),-\delta\right)\\
    &\leq \underbrace{\left|\nabla_\Theta \mathcal{L}_i(\Theta)\right|}_{\text{Fixed Value}}\underbrace{\left|\left|\delta\right|\right|}_{\text{Redundant Vector Norm}} \label{R_i_obj}
\end{align}

From Eq.~\ref{R_i_obj}, when we incorporate the norm of $\delta$ into the optimization objective, it optimizes the upper bound of the Replacement Gap $\mathcal{R}_i$. 
This is beneficial for the disentangled task vectors to closely approximate the performance of the original task vectors on their respective specific tasks.

\subsection{Solutions for Task Consistency Property}

\label{sec:proof}

In our paper, we define Task Consistency Property as the performance of the merged multi-task model on each individual task should approximate the performance obtained when only the corresponding task vector is integrated. 
This definition serves as a natural extension of Task Arithmetic Property~\cite{ta_ntk}, aligning more closely with the objectives of multi-task model merging and presenting a more tractable solution. The formal expression is as follows:

\noindent\textbf{Task Consistency Property}. \emph{Given coefficient sets $\{\lambda_i\}_{i=1}^K\subset \mathbb{R}$ and a set of task vectors $\mathcal{T} = \{\tau_i\}_{i=1}^{K}$ correspond with non-intersecting task-specific data supports $\mathcal{D}=\{\mathcal{D}_i\}_{i \in [K]}$, i.e., $\forall t,t^\prime$, if $t\neq t^\prime$ then $\mathcal{D}_t\cap\mathcal{D}_{t^\prime}=\varnothing$. We say a network $f$ satisfies the task consistency property around $\Theta$ with respect to $\mathcal{T}$ and $\mathcal{D}$ if:}
\begin{align} 
\forall \; \mathcal{D}_i\in \mathcal{D} \; , \; \mathcal{L}_i\left(\Theta+\sum_{j}^{K}\lambda_j\tau_j\right)=
    \mathcal{L}_i\left(\Theta+\lambda_i\tau_i\right),
    \label{eq:task_consistency}
\end{align}
where $\mathcal{L}_i(\cdot)$ denotes the loss function of task $i$. 
We fix the pre-trained model and solve for the task vector that satisfies task consistency property, which guides us in refining the task vector.
To simplified the Eq.~\ref{eq:task_consistency}, we define the \textbf{Merging Gap} $G_i$ as, 
\begin{align}
    G_{i} & = \mathcal{L}_i \left( \Theta  + \sum_{j}^K\lambda_j\tau_j \right)  - \mathcal{L}_i \left( \Theta + \lambda_i\tau_i \right).
    \label{G_i}
\end{align}

Based on the aforementioned formulation, the task consistency property can be reformulated as follows,
\begin{align}
    \forall\; i,\;G_i=0.
\end{align}

We apply Tylor expansion to $\mathcal{L}_i \left( \Theta  + \sum_{j}^K\lambda_j\tau_j \right)$ and $\mathcal{L}_i \left( \Theta + \lambda_i\tau_i \right)$ around $\Theta$.
We take $\mathcal{L}_i \left( \Theta  + \sum_{j}^K\lambda_j\tau_j \right)$ as an instance, it can be approximated as:
\begin{align}
    \mathcal{L}_i \left( \Theta  + \sum_{j}^K\lambda_j\tau_j \right) \approx \mathcal{L}_i(\Theta) + \left<\nabla_\Theta\mathcal{L}_i(\Theta),\sum_{j}^K\lambda_j \tau_j\right> \\
    + \frac{1}{2}\left(\sum_j^K\lambda_j\tau_j \right)^\top\mathbf{H}_i\left(\sum_j^K\lambda_j\tau_j \right),
\end{align}
where $\mathbf{H}_i$ is the Hessian of the loss $\mathcal{L}_i$ at $\Theta$.
Given that $\lambda_j$ typically falls within the range of $(-1, 1)$ and $\tau_j$ is generally quite small~\cite{dare}, the second-order terms can generally be approximated as zero under standard conditions. Eq.~\ref{G_i} can be simplified as,
\begin{align}
    G_{i} & = \mathcal{L}_i \left( \Theta  + \sum_{j}^K\lambda_j\tau_j \right)  - \mathcal{L}_i \left( \Theta + \lambda_i\tau_i \right) \\
&\approx\left( \mathcal{L}_i \left( \Theta  \right) + \left< \nabla_{\Theta} \mathcal{L}_i(\Theta),  \sum_{j}^K\lambda_j\tau_j \right> \right) \\
&-  \left( \mathcal{L}_i \left( \Theta  \right)
+ \left< \nabla_{\Theta} \mathcal{L}_i(\Theta), \lambda_i\tau_i  \right> \right) \\
&= \left< \nabla_{\Theta} \mathcal{L}_i(\Theta), \sum_{j}^K\lambda_j\tau_j \right>
- \left< \nabla_{\Theta} \mathcal{L}_i(\Theta), \lambda_i\tau_i  \right> \\
&= \left< \nabla_{\Theta} \mathcal{L}_i(\Theta), \lambda_i\tau_i + \sum_{j\neq i}^K\lambda_j\tau_j \right>
- \left< \nabla_{\Theta} \mathcal{L}_i(\Theta), \lambda_i\tau_i  \right> \\
&= \left< \nabla_{\Theta} \mathcal{L}_i(\Theta), \sum_{j\neq i}^K\lambda_j\tau_j \right>\\
&= \sum_{j\neq i}^K \left< \nabla_{\Theta} \mathcal{L}_i(\Theta), \lambda_j\tau_j \right>.
\end{align}

Obtaining the original data for gradient calculation in a pre-trained model for a specific task can often be challenging due to accessibility issues. 
An effective alternative is to substitute this gradient with the task vector, as the task vector can be conceptualized as the accumulation of gradients.
Therefore, Eq.~\ref{G_i} can be described as,
\begin{align}
    G_i &= \sum_{j\neq i}^K\left< k_i\tau_i, \lambda_j\tau_j \right>\\
    &= k_i\sum_{j\neq i}^K\left< \tau_i, \lambda_j\tau_j \right>,
    \label{G_i_final}
\end{align}
where $k_i<0$.
Since $\lambda_j$ is typically a hyperparameter, directly solving the previously mentioned equation is challenging.
Consequently, our objective is to identify a specific solution for this equation. 
As demonstrated in Eq.~\ref{G_i_final}, Eq.~\ref{eq:task_consistency} will be satisfied when all task vectors are orthogonal to each other,
\begin{align}
   \forall \; i\ne j, \; \cos \left ( \tau_i,\tau_j  \right ) = 0  \Rightarrow \forall i, \; G_i = 0.
\end{align}

\begin{table*}[ht]
\centering
\caption{Multi-task performance when merging ViT-B/16 models on eight tasks. The best performance across different merging methods is denoted in bold.}
\label{tab:performance_vitbase16} 
\resizebox{\linewidth}{!}{  
\begin{tabular}{l|cccccccc|cc}
\toprule
\textbf{Method}  &  \textbf{SUN397}  &  \textbf{Cars}  &  \textbf{RESISC45}  &  \textbf{EuroSAT}  &  \textbf{SVHN}  &  \textbf{GTSRB}  &  \textbf{MNIST}  &  \textbf{DTD}  & \textbf{Avg Acc}  \\
\midrule
\multicolumn{10}{c}{\emph{Non-merging Methods}} \\
{Pretrained}  &  63.8 & 64.6 & 65.7 & 54.5 & 52.0 & 43.3 & 51.7 & 45.1 & 55.0   \\
{Individual}  &  81.8 & 86.8 & 96.9 & 99.7 & 97.8 & 99.1 & 99.7 & 82.0 & 92.9\\
\midrule
\multicolumn{10}{c}{\emph{Training-free Methods}} \\
{Weight Averaging} & 67.7 & 70.0 & 75.3 & 79.5 & 74.9 & 60.1 & 94.4 & 43.8 & 70.7 \\
{Fisher Merging}   & 68.5 & 69.9 & 75.2 & 80.4 & 73.2 & 61.2 & 94.5 & 50.7 & 71.7\\
{RegMean}          & 69.1 & 71.6 & 77.6 & \textbf{88.8} & 83.7 & 70.2 & 96.9 & 54.6 & 76.6\\
Task Arithmetic    & 61.1 & 65.9 & 74.0 & 76.2 & 88.0 & 73.9 & 98.4 & 53.0 & 73.8\\
{Ties-Merging}     & 69.1 & 72.5 & 80.5 & 84.0 & 85.0 & 71.5 & 98.1 & 54.9 & 77.0\\
Consensus Merging  & \textbf{69.8} & 71.4 & \textbf{80.8} & 86.5 & 88.0 & 71.1 & 98.4 & 57.0 & 77.9\\
\textbf{AWD Task Arithmetic} (Ours)& 67.8 & \textbf{72.7} & 78.7 & 88.5 & \textbf{90.9} & \textbf{83.6} & \textbf{98.9} & \textbf{57.1} & \textbf{79.8}\\
\midrule
\multicolumn{10}{c}{\emph{Test-time Adaption Methods}} \\

{TW AdaMerging}   & 64.4 & 64.2 & 75.4 & 86.7 & 86.3 & 86.7 & 97.6 & 46.9 & 76.0\\
{LW AdaMerging}   & 70.2 & 80.7 & 81.6 & 94.8 & 91.6 & 95.8 & 98.5 & 66.2 & 84.9\\
{Representation Surgery} & 68.3 & 72.3 & \textbf{88.7} & \textbf{97.7} & 91.0 & 89.5 & \textbf{98.9} & \textbf{72.9} & 84.9\\
\textbf{AWD AdaMerging} (Ours) & \textbf{71.5} & \textbf{81.9} & 84.3 & 94.6 & \textbf{93.3} & \textbf{96.8} & 98.6 & 72.6 & \textbf{86.7}\\ %

\bottomrule
\end{tabular}
}
\end{table*}

\section{Experimental Settings}

\begin{table}[h]
\small
\centering
\caption{{Computational time and GPU memory requirements for solving redundant vector across eight vision tasks over 1000 solution steps were assessed on NVIDIA A100 40GB.}}
\label{tab:trainning_cost_Appendix} 
\resizebox{0.75\linewidth}{!}{  
\begin{tabular}{l|cc}
\toprule
Model  & Solving Time & GPU Memory \\
\midrule
ViT-B/32 & 2min22s & 3.76GB \\
ViT-L/14 & 7min43s & 5.04GB \\
\bottomrule
\end{tabular}
}
\end{table}

\begin{table*}[ht]
\centering
\caption{Multi-task performance when merging ViT-B/32 models on 8-task vision benchmark, where Individual Reconstructed refers to task-specific models that are reconstructed by combining disentangled task vectors with the pre-trained model respectively. 
By utilizing the orthogonality optimization objective, we can derive disentangled task vectors with increased orthogonality. Conversely, by employing a reverse optimization objective, we can obtain disentangled task vectors with decreased orthogonality.
}
\label{tab:vit_32_neg} 
\resizebox{0.95\linewidth}{!}{  
\begin{tabular}{l|cccccccc|cc}
\toprule
\textbf{Method}  &  \textbf{SUN397}  &  \textbf{Cars}  &  \textbf{RESISC45}  &  \textbf{EuroSAT}  &  \textbf{SVHN}  &  \textbf{GTSRB}  &  \textbf{MNIST}  &  \textbf{DTD}  & \textbf{Avg Acc}  \\
\midrule
{Pretrained}  &  {62.3}  &  {59.7}  &  {60.7}  &  {45.5}  &  {31.4}  &  {32.6}  &  {48.5}  &  {43.8} & {48.0}   \\
{Individual}  &  79.2  &  77.7  &  96.1  &  99.7  &  97.5  &  98.7  &  99.7  &  79.4 & 90.8   \\
Task Arithmetic   &55.2 &54.9 &66.7 &78.9 &80.2 & 69.7 &97.3 &50.4 & 69.1 \\

\midrule
\multicolumn{10}{c}{\emph{Increased Orthogonality (\textbf{Ours})}} \\
{Individual Reconstructed} & 79.3 & 77.6 & 96.1 & 99.8 & 97.4 & 98.8 & 99.7 & 78.8 & 90.9 \\
{AWD Task Arithmetic}  & 63.5 & 61.9  & 72.6 & {84.9} & 85.1 & {79.1} & 98.1 & 56.7 & {75.2}\\ %

\midrule
\multicolumn{10}{c}{\emph{Decreased Orthogonality (\textbf{Reversed}})} \\
{Individual Reconstructed} & 78.3 & 76.6 & 95.6 & 99.6 & 97.4 & 98.8  & 99.7  & 78.6 & 90.6 \\
{AWD Task Arithmetic}  & 2.7 & 1.1 & 19.3 & 42.1 & 60.1 & 20.1 & 92.6 & 16.7 & 31.8\\
\bottomrule
\end{tabular}
}
\end{table*}

\begin{table*}[t]
\centering
\caption{Multi-task performance when merging ViT-L/14 models on 8-task vision benchmark, where Individual Reconstructed refers to task-specific models that are reconstructed by combining disentangled task vectors with the pre-trained model respectively.}
\label{tab:vit_14_neg} 
\resizebox{0.95\linewidth}{!}{  
\begin{tabular}{l|cccccccc|cc}
\toprule
\textbf{Method}  &  \textbf{SUN397}  &  \textbf{Cars}  &  \textbf{RESISC45}  &  \textbf{EuroSAT}  &  \textbf{SVHN}  &  \textbf{GTSRB}  &  \textbf{MNIST}  &  \textbf{DTD}  & \textbf{Avg Acc}  \\
\midrule
 {Pretrained}  & {66.8}  & {77.7}  & {71.0}  &   {59.9}  & {58.4}  &  {50.5}  & {76.3}  &   {55.3} &  {64.5}\\
{Individual}   &  82.3  &  92.4  &  97.4  &  100  &  98.1  &  99.2  &  99.7  &  84.1  & 94.2   \\
{Task Arithmetic}   &73.9  &82.1 &86.6 &94.1  &87.9  &86.7  &98.9  &65.6   &84.5 \\
\midrule
\multicolumn{10}{c}{\emph{Increased Orthogonality (\textbf{Ours})}} \\
{Individual Reconstructed} & 84.8 & 92.4 & 97.3 & 99.7 & 98.1 & 99.3 & 99.7 & 84.3 & 94.4\\
{AWD Task Arithmetic} & 76.2 & 85.4 & 88.7 & 96.1 & {92.4} & {92.3} & {99.3} & 69.4 & {87.5}\\
\midrule
\multicolumn{10}{c}{\emph{Decreased Orthogonality (\textbf{Reversed}})} \\
{Individual Reconstructed} & 84.6 & 92.3 & 97.4 & 99.7 & 98.2  & 99.3 & 99.7  & 84.4 & 94.4\\
{AWD Task Arithmetic}  & 36.0 & 31.5 & 48.5 & 54.9 & 83.2 & 54.3 & 97.6 & 37.9 & 55.5 \\
\bottomrule
\end{tabular}
}
\end{table*}

\subsection{Calculate Resources and Environment.}
All of our experiments were conducted on NVIDIA A100 40GB and NVIDIA H800 80GB. 
Due to the specific configuration of AdaMerging, we used the NVIDIA H800 80GB for the ViT-L/14 variant, while other experiments were conducted on the NVIDIA A100 40GB.
The software environment for our experiments comprised Python 3.10, PyTorch 2.4.0, and the CUDA 11.8 toolkit for the A100, and 12.1 for the H800. 
As illustrated in Table~\ref{tab:trainning_cost_Appendix}, our methodology incurs minimal computational overhead across different model versions, demonstrating near-universal scalability on devices equipped with contemporary GPUs.

\subsection{Checkpoints for Merging.}
For the vision models, the checkpoints we utilized are sourced from~\citet{task_arithmetic}, which were obtained by fine-tuning the CLIP model~\cite{clip} using the AdamW optimizer with a weight decay of 0.1. The parameters involved include a batch size of 128 and a learning rate of 1e-5, employing a cosine annealing learning rate schedule over 2000 iterations, which includes 200 warm-up steps.
For the language models, we utilize RoBERTa-Base and RoBERTa-Large as the pretrained backbones. Our checkpoints are derived from~\citet{Lu2024TwinMerging}. These checkpoints underwent 10 epochs of fine-tuning on the RoBERTa model for each dataset, with selected hyperparameters including a batch size of 64 and a learning rate of $1e-5$.

\subsection{Hyper-parameter tuning.}
To balance the relationship between orthogonality and invariance, we tune the hyper-parameter Constraints Coefficient $\alpha$ over a range of $\{1e-2, 1e-3, 1e-4, 1e-5, 1e-6\}$. 
For {AWD Task Arithmetic:} We follow~\citet{task_arithmetic}, and we use a single scaling factor $\lambda$ to scale the sum of the disentangled task vectors for the model merging. The scaling factor $\lambda$ is tuned over a range of $\{0.3, 0.4, ..., 1.0\}$. 
For AWD AdaMerging: we use 0.3 or 0.4 to initialize each $\lambda_{ij}$.
The best $\alpha$ and $\lambda$ are selected based on the performance of the validation set averaged on all tasks.

\begin{table*}[ht]
\centering
\caption{Multi-task performance when merging RoBERTa-Base models on 8-task GLUE benchmark. We report normalized score~\cite{Lu2024TwinMerging}. The best and second-best performing model combination methods for each task are indicated in bold and underlined, respectively. }
\label{tab:performance_roberta_base} 
\resizebox{0.87\linewidth}{!}{  
\begin{tabular}{l|cccccccc|cc}
\toprule
\textbf{Method} & \textbf{CoLA} & \textbf{SST-2} & \textbf{MRPC} & \textbf{STS-B} & \textbf{QQP} & \textbf{QNLI} & \textbf{MNLI} & \textbf{RTE} & \textbf{Avg}  \\

\midrule
Pretrained & 0.0 & 53.8 & 85.0 & 4.0 & 37.5 & 53.1 & 37.1 & 71.2 & 41.7 \\
{Individual}  & 100.0 & 100.0 & 100.0 & 100.0 & 100.0 & 100.0 & 100.0 & 100.0 & 100.0 \\
\midrule
Weight Averaging & 0.0  & 59.2 & 85.8 & \textbf{47.0} & 45.4 & 63.9 & 48.0 & 71.2 & 52.6 \\
Task Arithmetic & 8.4 & 88.3 & \textbf{89.6} & 32.8 & 82.0 & 85.4 & 75.5 & \underline{80.4} & \underline{67.8} \\
{Ties-Merging} & \textbf{31.8} & \underline{88.9} & 86.2 & 10.9 & 61.1 & \underline{85.9} & \underline{83.0} & 69.6 & 64.7 \\
Task Arithmetic (w/ DARE) & 0.0 & 88.1 & 86.6 & 30.2 & \underline{84.3} & 79.1 & 64.0 & 77.2 & 63.7 \\
{Ties-Merging} (w/ DARE) & \underline{11.8} & \textbf{95.5} & 85.8 & 9.4 & \textbf{86.8} & \textbf{88.7} & \textbf{83.1} & 63.6 & 65.6 \\
\textbf{AWD Task Arithmetic} (Ours)& \underline{11.8} & 88.3 & \underline{89.1} & \underline{33.2} & 80.8 & 85.7 & 76.1 & \textbf{81.5} & \textbf{68.3}\\
\bottomrule
\end{tabular}
}
\end{table*}

\begin{table*}[ht]
\centering
\caption{Multi-task performance when merging RoBERTa-Large models on 8-task GLUE benchmark. We report normalized score~\cite{Lu2024TwinMerging}. The best and second-best performing model combination methods for each task are indicated in bold and underlined, respectively. 
}
\label{tab:performance_roberta_large} 
\resizebox{0.87\linewidth}{!}{  
\begin{tabular}{l|cccccccc|cc}
\toprule
\textbf{Method} & \textbf{CoLA} & \textbf{SST-2} & \textbf{MRPC} & \textbf{STS-B} & \textbf{QQP} & \textbf{QNLI} & \textbf{MNLI} & \textbf{RTE} & \textbf{Avg}  \\

\midrule
Pre-trained & 0.0 & 51.5 & 40.9 & 20.9 & 36.4 & 56.0 & 37.6 & 62.4 & 38.2\\
{Individual}  & 100.0 & 100.0 & 100.0 & 100.0 & 100.0 & 100.0 & 100.0 & 100.0 & 100.0 \\
\midrule
Weight Averaging & 7.4  & 55.1 & 84.2 & 46.3 & 56.7 & 73.8 & 35.8 & 66.7 & 53.3\\
Task Arithmetic & 7.4 & 86.1 & \textbf{86.8} & \underline{78.0} & 90.7 & 77.0 & 73.3 & 67.6 & 70.9 \\
{Ties-Merging} & \textbf{42.7} & 78.1 & 85.2 & 51.7 & 89.9 & 81.9 & \textbf{79.7} & \textbf{70.0} & 72.4 \\ %
Task Arithmetic (w/ DARE) & 4.1 & 85.2 & 85.8 & 71.6 & 91.3 & 85.6 & 75.2 & 68.1 & 70.9\\
{Ties-Merging} (w/ DARE) & 2.9 & \underline{90.4} & \textbf{86.8} & 75.4 & \textbf{92.4} & \underline{86.4} & \underline{79.0} & \underline{69.1} & \underline{72.8}\\ 
\textbf{AWD Task Arithmetic} (Ours)& \underline{10.4} & \textbf{90.8} & \underline{86.5} & \textbf{87.1} & \underline{91.8} & \textbf{87.1} & 78.1 & 63.8 & \textbf{74.5}\\ %
\bottomrule
\end{tabular}
}
\end{table*}

\section{Additional Experimentals}

\subsection{Performance on ViT-B/16}

Table~\ref{tab:performance_vitbase16} presents the performance of various merging methods applied to ViT-B/16. Our approach achieves a 6.0\% improvement over Task Arithmetic and a 1.8\% gain compared to AdaMerging on corresponding settings. These results strongly indicate that our method effectively mitigates conflicts between task vectors, further demonstrating the effectiveness of our AWD in multi-task model merging.

\begin{table*}[ht]
\centering
\caption{Generalization results on two unseen tasks when merging ViT-B/16 models on six tasks.}
\label{tab:ood_generalize_vitb16_appendix} 
\resizebox{0.95\linewidth}{!}{  
\begin{tabular}{l|ccccccc|cccc}
\toprule
& \multicolumn{7}{c|}{\textbf{Seen Tasks}} & \multicolumn{3}{c}{\textbf{Unseen Tasks}} \\
\midrule \midrule
{Method}   &  {SUN397}  &  {Cars}  &  {RESISC45}   &  {DTD} &  {SVHN}  &  {GTSRB}  & \textbf{Avg Acc}  &  {EuroSAT} &  {MNIST}   & \textbf{Avg Acc} \\
\midrule
{Task Arithmetic}    & 68.1 & 73.0 & 81.6 & 59.1 & 89.1 & 83.8 & 75.8 & 43.9 & 87.5 & 65.7\\
\textbf{AWD Task Arithmetic} (Ours)    & 70.9 & 76.3 & 85.1 & 61.9 & 91.7 & 88.7 & 79.1 & 44.6 & 87.7 & 66.2\\
\midrule \midrule
{Method}    &  {SUN397}  &  {Cars} &  {GTSRB}  &  {EuroSAT}    &  {DTD}  &  {MNIST}  & \textbf{Avg Acc}  &  {RESISC45} &  {SVHN}    & \textbf{Avg Acc} \\
\midrule
{Task Arithmetic}    & 69.0 & 73.8 & 81.1 & 87.6 & 58.2 & 98.4 & 78.0 & 56.0 & 67.7 & 61.8\\
\textbf{AWD Task Arithmetic} (Ours)  & 72.3 & 77.0 &  85.4 & 93.5 & 76.7 & 98.5 & 81.1 & 58.2 & 68.6 & 63.4\\
\bottomrule
\end{tabular}
}
\end{table*}

\begin{table*}[ht]
\centering
\caption{Comparison of performance across multiple task setups using various methods. Red text indicates the performance improvements compared to Task Arithmetic for each configuration.}
\label{tab:performance_comparision_different_task_numbers} 
\resizebox{0.8\linewidth}{!}{  
\begin{tabular}{l|cccccccc}
\toprule
\textbf{Method} &  \textbf{2-Tasks}    &  \textbf{3-Tasks}  &  \textbf{4-Tasks}  &  \textbf{5-Tasks}  &  \textbf{6-Tasks}  & \textbf{7-Tasks} & \textbf{8-Tasks}\\
\midrule
Individual             & 93.0 & 90.6 & 92.5 & 90.1 & 91.5 & 91.4 & 90.6\\
\midrule
Task Arithmetic        & 91.8 & 84.1 & 82.3 & 77.1 & 75.7 & 72.6 & 69.1\\
{Ties-Merging}         & $92.6_{\textcolor{red}{\bigtriangleup 0.8}}$ & $85.6_{\textcolor{red}{\bigtriangleup 1.5}}$ & $83.9_{\textcolor{red}{\bigtriangleup 1.6}}$ & $79.2_{\textcolor{red}{\bigtriangleup 2.1}}$& $77.6_{\textcolor{red}{\bigtriangleup 1.9}}$ & $75.1_{\textcolor{red}{\bigtriangleup 2.5}}$ & $72.4_{\textcolor{red}{\bigtriangleup 3.3}}$\\
\textbf{AWD Task Arithmetic} (Ours)   & $92.4_{\textcolor{red}{\bigtriangleup 0.6}}$ & $85.7_{\textcolor{red}{\bigtriangleup 1.6}}$ & $85.1_{\textcolor{red}{\bigtriangleup 2.8}}$ & $80.6_{\textcolor{red}{\bigtriangleup 3.5}}$ & $79.7_{\textcolor{red}{\bigtriangleup 4.0}}$ & $77.1_{\textcolor{red}{\bigtriangleup 4.5}}$ & $75.2_{\textcolor{red}{\bigtriangleup 6.1}}$ \\
\bottomrule
\end{tabular}
}
\end{table*}

\begin{figure}[t]
\centering
\begin{subfigure}{0.48\linewidth}
    \includegraphics[width=\linewidth]{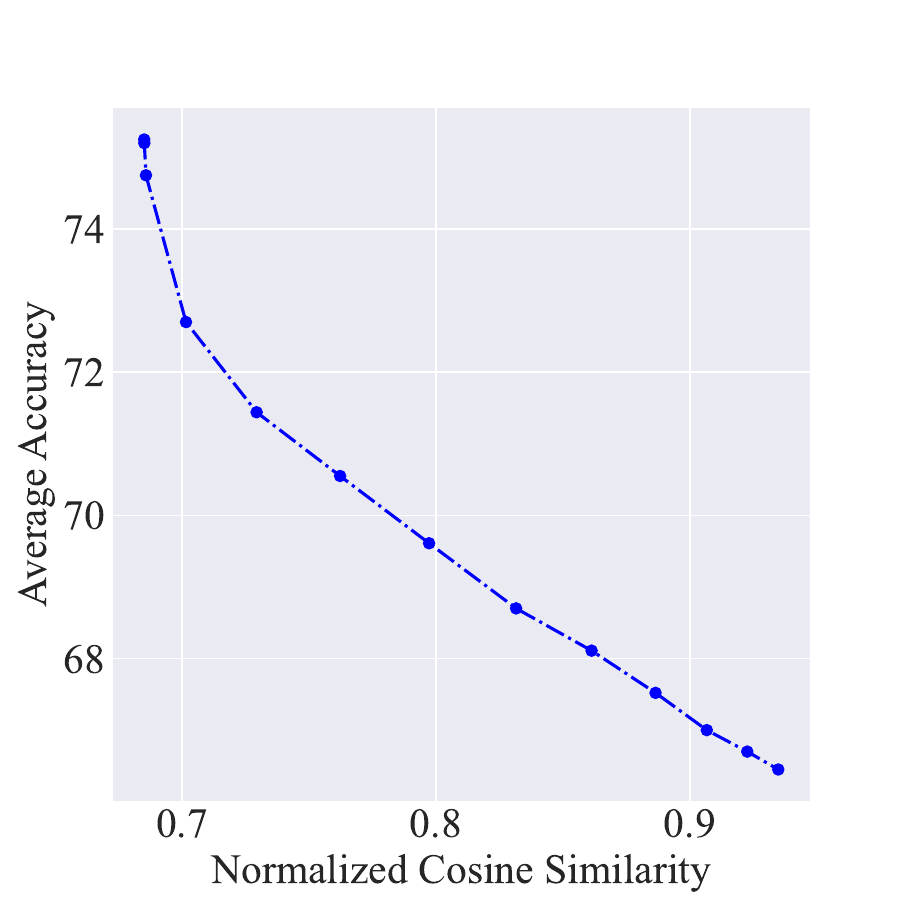}
    \subcaption{ViT-B/32}
    \label{cosine_perf_vit_b_32}
\end{subfigure}
\hfill
\begin{subfigure}{0.48\linewidth}
    \includegraphics[width=\linewidth]{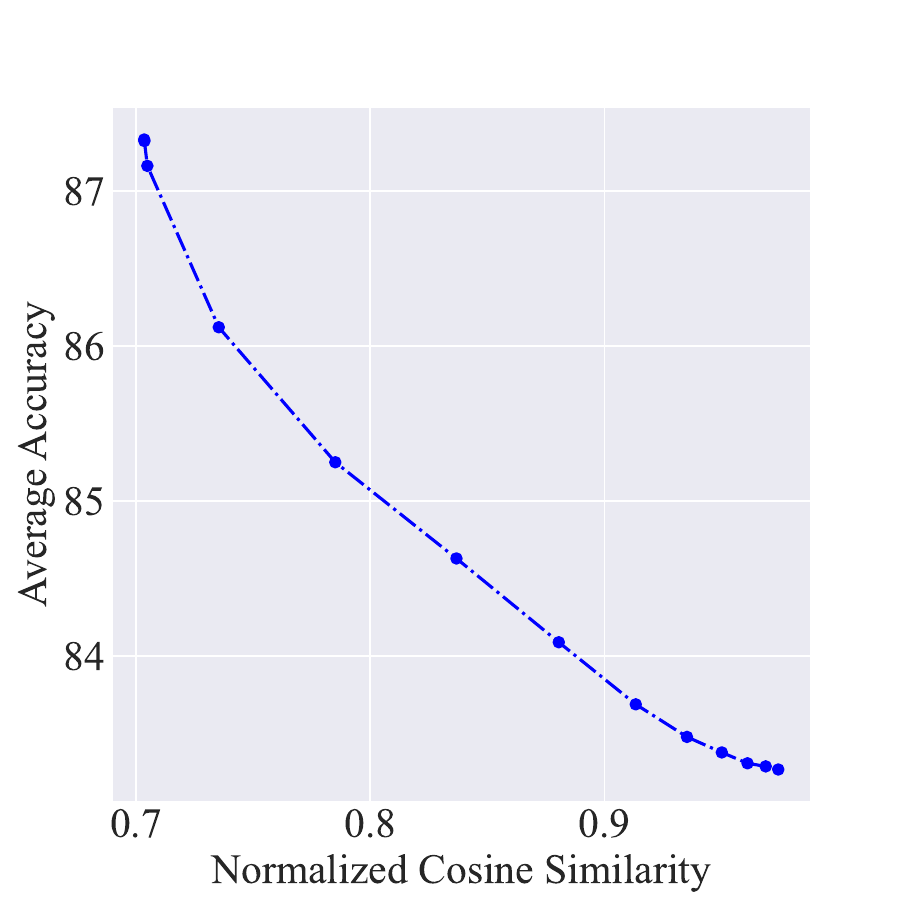}
    \subcaption{ViT-L/14}
    \label{cosine_perf_vit_l_14}
\end{subfigure}
\caption{Correlation between normalized cosine similarity between disentangled task vectors and merged model performance. The normalized cosine score is calculated by normalizing the cosine similarity between the disentangled task vectors with respect to that of the original task vectors.}
\label{robust_analysis}
\end{figure}

\subsection{More Discussions on Orthogonality}

To further validate the effectiveness of our approximate orthogonal solutions, we conducted a detailed analysis of the relationship between the cosine similarity between task vectors and the performance of the merged model. Given the fixed $\lambda$, by tuning the value of $\alpha$, we ensured that the disentangled task vector could proficiently restore the performance of the specific task associated with the original task vector, concurrently adjusting the orthogonality among the disentangled task vectors.

Fig.~\ref{cosine_perf_vit_b_32} and Fig.~\ref{cosine_perf_vit_l_14} depict the correlation between normalized cosine similarity of the task vectors and the performance of the merged model on ViT-B/32 and ViT-L/14, respectively. 
We observe that as the normalized cosine similarity between disentangled task vectors decreases, the performance of the merged model correspondingly improves. Based on these findings, we can infer that enhancing the orthogonality among disentangled task vectors is beneficial in mitigating interference between them, thereby achieving a more effective fusion.
This further bolsters our strategy of considering the orthogonality of the disentangled task vectors as an optimization objective.

Furthermore, we investigate the impact of reducing orthogonality on model merging by reversing the orthogonality optimization objective, as shown in Table~\ref{tab:vit_32_neg} and Table~\ref{tab:vit_14_neg}. Under different optimization objectives, we ensure that the disentangled task vectors can still recover the performance of the original task vectors. 
Compared to task arithmetic, by reversing the optimization objective, our AWD exhibits a performance decrease of 37.3\% on ViT-B/32 and 29.0\% on ViT-L/14.
This implies that reducing the orthogonality among task vectors could intensify the interference between them, potentially undermining the model's capacity for effective task integration.

\subsection{Generalization in Language Models}

\noindent\textbf{Metrics}:
Following~\citet{Lu2024TwinMerging}, we utilize the normalized scores as the assessment criterion. Similarly, we define the individually fine-tuned models as the upper performance bounds and the pre-trained models as the lower bounds. The Normalized Score is described as follows:
\begin{align}
    \text{Normalized Score} = \frac{1}{K} \sum_{i}^{K} \frac{\mathop{\text{Score}}\limits_{x\in\mathcal{D}_i} [f(x; \widehat{\Theta}]}{ \mathop{\text{Score}}\limits_{x\in\mathcal{D}_i} [f(x; \Theta_i^\star)]}
\end{align}

\noindent\textbf{Detailed Results}:
In the main paper, we demonstrated the effectiveness of our approach on the RoBERTa language models. As illustrated in Table~\ref{tab:performance_roberta_base} and Table~\ref{tab:performance_roberta_large}, we further present detailed results of our method across various tasks.

\subsection{Generalization on Unseen Tasks}

As shown in Table \ref{tab:ood_generalize_vitb16_appendix}, we conducted a comprehensive evaluation of AWD on both in-domain and out-of-domain tasks. Under the ViT-B/16 configuration, AWD demonstrated exceptional generalization performance. For in-domain tasks, our method achieved an average improvement of 3.3\% across different settings. In cases where the corresponding task vectors were not merged (i.e., unseen tasks), our AWD method improved the average accuracy by 0.5\% over the task arithmetic method on the EuroSAT and MNIST tasks; on the unseen tasks of RESISC45 and SVHN, the average accuracy was increased by 1.6\%.

\subsection{Empirical Upper Bound of AWD Task Arithmetic}

\begin{figure}[t]
\centering
\begin{subfigure}{0.48\linewidth}
    \includegraphics[width=\linewidth]{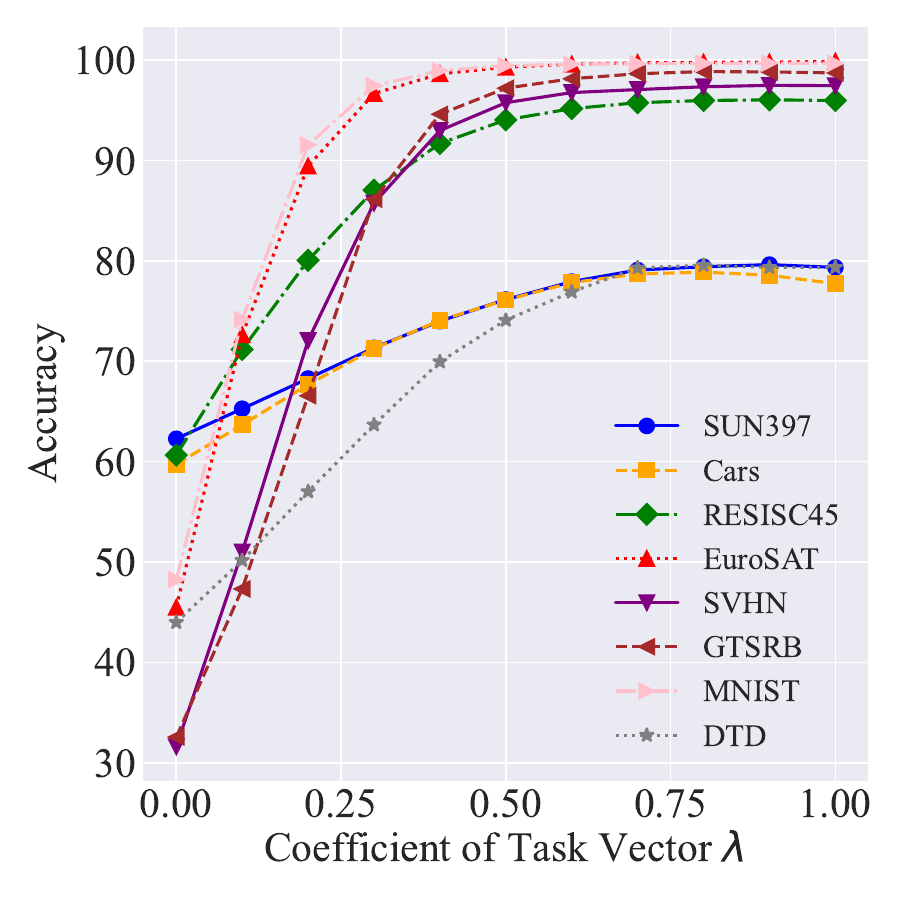}
    \subcaption{ViT-B/32}
    \label{individual_perf_vit_b_32}
\end{subfigure}
\hfill
\begin{subfigure}{0.48\linewidth}
    \includegraphics[width=\linewidth]{figures/individual_vit_32.pdf}
    \subcaption{ViT-L/14}
    \label{individual_perf_vit_l_14}
\end{subfigure}
\caption{Performance of the integrated model $\Theta+\lambda\tau_i$ on corresponding task $i$ across various values of $\lambda$. The integrated model $\Theta+\lambda\tau_i$ denotes as the combination of pre-trained models with individual task vector.}
\label{robust_analysis}
\end{figure}

For conventional training-free methods based on task vectors, there typically exists an adjustable coefficient $\lambda$. 
A disproportionately small $\lambda$ may result in suboptimal performance of the task vector on its corresponding task, while an excessively large $\lambda$ may amplify interference between tasks. 
Consequently, this coefficient plays a pivotal role in determining the performance of the merged model.
According to Eq.~\ref{eq:task_consistency}, it can be seen that, the approximate upper bound of performance of our AWD Task Arithmetic on a specific task $i$ can be represented as the performance of the integrated model $\Theta+\lambda \tau_i$.
Here, $\Theta+\lambda \tau_i$ represents the amalgamation of the pretrained model with the individual task vector $\tau_i$, which is proportionally scaled by $\lambda$.
Therefore, we further conducted experiments on the performance under different $\lambda$ coefficients. 
As shown in Fig.~\ref{individual_perf_vit_b_32} and Fig.~\ref{individual_perf_vit_l_14}, when $\lambda>0.3$ for each combination of $\Theta+\lambda\tau_i$, its performance is close to that of the fully fine-tuned model $\Theta_i^\star=\Theta + \tau_i$. 
Moreover, as $\lambda$ continues to increase, the rate of performance improvement sharply decreases. 
Notably, when $\lambda$ is greater than 0.3, its performance has already surpassed the optimal merging method significantly. 
These results indicate that the approximate upper bound of our AWD Task Arithmetic is reasonable when $\lambda\geq 0.3$.

\subsection{Performance across Different Task Numbers}

In Table~\ref{tab:performance_comparision_different_task_numbers}, we further evaluated our method across varying numbers of tasks. Specifically, when merging $T$ tasks, there are a total of $\binom{8}{T}$ possible combinations. Therefore, in our experiment, we sampled 8 distinct combinations for each value of $T$ and performed a hyperparameter search within the range $[0, 1]$ with a step size of $0.1$ for each combination.
It is evident that as the number of tasks increases, our method demonstrates enhanced robustness against interference.
For a more detailed analysis, please refer to Section 4.5 of the main paper.

The sampled task combinations are listed as follows:
\begin{itemize}
\item 2-Tasks: [[DTD, GTSRB], [GTSRB, SVHN], [GTSRB, SUN397], [SVHN, SUN397], [SVHN, GTSRB], [SVHN, EuroSAT], [SVHN, MNIST], [MNIST, Cars]]
\item 3-Tasks: [[MNIST, SVHN, SUN397], [MNIST, SUN397, SVHN], [DTD, GTSRB, SUN397], [GTSRB, EuroSAT, Cars], [Cars, GTSRB, DTD], [MNIST, RESISC45, SVHN], [SVHN, MNIST, DTD], [RESISC45, SUN397, EuroSAT]]
\item 4-Tasks: [EuroSAT, SVHN, Cars, SUN397], [MNIST, RESISC45, EuroSAT, GTSRB], [EuroSAT, Cars, RESISC45, MNIST], [DTD, SUN397, GTSRB, EuroSAT], [SUN397, EuroSAT, Cars, RESISC45], [RESISC45, MNIST, GTSRB, SUN397], [RESISC45, SVHN, GTSRB, MNIST], [SVHN, GTSRB, Cars, RESISC45]
\item 5-Tasks: [[DTD, SVHN, GTSRB, SUN397, EuroSAT], [DTD, GTSRB, MNIST, RESISC45, SUN397], [SVHN, MNIST, GTSRB, RESISC45, Cars], [DTD, EuroSAT, Cars, MNIST, RESISC45], [EuroSAT, GTSRB, MNIST, Cars, RESISC45], [MNIST, Cars, SUN397, DTD, SVHN], [MNIST, SUN397, RESISC45, SVHN, DTD], [SVHN, DTD, Cars, SUN397, MNIST]]
\item 6-Tasks: [[GTSRB, RESISC45, DTD, MNIST, SVHN, SUN397], [EuroSAT, GTSRB, Cars, MNIST, DTD, RESISC45], [MNIST, SUN397, SVHN, RESISC45, EuroSAT, DTD], [DTD, MNIST, RESISC45, SVHN, GTSRB, SUN397], [SVHN, RESISC45, EuroSAT, MNIST, GTSRB, DTD], [MNIST, DTD, EuroSAT, Cars, SUN397, GTSRB], [DTD, Cars, SVHN, SUN397, EuroSAT, MNIST], [SVHN, SUN397, RESISC45, GTSRB, EuroSAT, MNIST]]
\item 7-Tasks: [[GTSRB, MNIST, Cars, RESISC45, SVHN, DTD, EuroSAT], [Cars, GTSRB, MNIST, SVHN, SUN397, EuroSAT, RESISC45], [Cars, MNIST, SUN397, DTD, EuroSAT, GTSRB, SVHN], [GTSRB, SUN397, EuroSAT, Cars, RESISC45, DTD, MNIST], [SVHN, Cars, GTSRB, MNIST, SUN397, EuroSAT, DTD], [EuroSAT, GTSRB, DTD, RESISC45, SVHN, MNIST, SUN397], [MNIST, SVHN, GTSRB, RESISC45, EuroSAT, DTD, Cars], [EuroSAT, MNIST, GTSRB, DTD, RESISC45, SVHN, SUN397]]
\end{itemize}

\end{document}